\documentclass[10pt,twocolumn,letterpaper]{article}

\usepackage{cvpr}              %
\usepackage[dvipsnames]{xcolor}
\usepackage[export]{adjustbox}
\usepackage{arydshln}
\usepackage{makecell}
\usepackage[hang,flushmargin]{footmisc}
\newcommand{\plusminus}[0]{\scalebox{0.6}[1.0]{$\pm$}}
\usepackage[accsupp]{axessibility}

\newcommand{\nohttpurl}[1]{%
    \IfBeginWith{#1}{http://}{%
        \href{#1}{\nolinkurl{\StrSubstitute{#1}{http://}{}}}%
    }{%
        \IfBeginWith{#1}{https://}{%
            \href{#1}{\nolinkurl{\StrSubstitute{#1}{https://}{}}}%
        }{%
            \href{http://#1}{\nolinkurl{#1}}%
        }%
    }%
}
\newcommand\extrafootertext[1]{%
    \bgroup
    \renewcommand\thefootnote{\fnsymbol{footnote}}%
    \renewcommand\thempfootnote{\fnsymbol{mpfootnote}}%
    \footnotetext[0]{#1}%
    \egroup
}
\usepackage{multirow} 
\usepackage{tabularray}
\usepackage[ruled,vlined]{algorithm2e}

\newcommand{\fakesection}[1]{%
  \par\refstepcounter{section}%
  \sectionmark{#1}%
  \addcontentsline{toc}{section}{\protect\numberline{\thesection}#1}%
}

\newcommand{\tableimg}[1]{
    \hspace{-8pt}
    \includegraphics[width=0.075\textwidth,valign=c]{#1}
    \hspace{-8pt}
}

\input{utils/custom}

\definecolor{cvprblue}{rgb}{0.21,0.49,0.74}
\usepackage[pagebackref,breaklinks,colorlinks,citecolor=cvprblue]{hyperref}

\def\paperID{12450} %
\def\confName{CVPR}
\def\confYear{2024}

\title{Joint-Task Regularization for Partially Labeled Multi-Task Learning}

\begin{document}
\author{
\vspace{-2em}\\
Kento Nishi* \qquad
Junsik Kim* \qquad
Wanhua Li \qquad 
Hanspeter Pfister\\
Harvard University\\
{\tt\small kentonishi@college.harvard.edu, \{jskim, wanhua, pfister\}@seas.harvard.edu}\\
\vspace{-2em}
}
\maketitle

\extrafootertext{*Equal contribution.}
\extrafootertext{Code is available at \nohttpurl{github.com/KentoNishi/JTR-CVPR-2024}.}

\begin{abstract}
    Multi-task learning has become increasingly popular in the machine learning field, but its practicality is hindered by the need for large, labeled datasets. Most multi-task learning methods depend on fully labeled datasets wherein each input example is accompanied by ground-truth labels for all target tasks. Unfortunately, curating such datasets can be prohibitively expensive and impractical, especially for dense prediction tasks which require per-pixel labels for each image. With this in mind, we propose Joint-Task Regularization (JTR), an intuitive technique which leverages cross-task relations to simultaneously regularize all tasks in a single joint-task latent space to improve learning when data is not fully labeled for all tasks. 
    JTR stands out from existing approaches in that it regularizes all tasks jointly rather than separately in pairs---therefore, it achieves linear complexity relative to the number of tasks while previous methods scale quadratically. 
    To demonstrate the validity of our approach, we extensively benchmark our method across a wide variety of partially labeled scenarios based on NYU-v2, Cityscapes, and Taskonomy. 
\end{abstract}    
\section{Introduction}
\label{sec:intro}
\begin{figure}[t]
    \centering
    \includegraphics[width=0.40\textwidth]{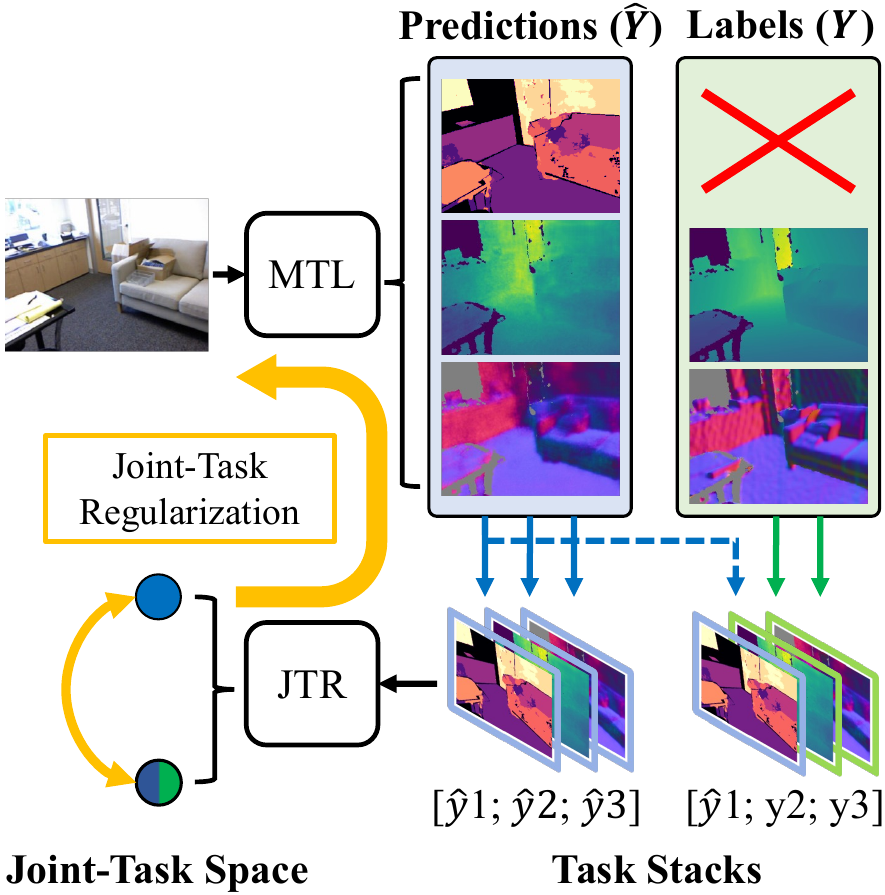}
    \caption{
        An overview of Joint-Task Regularization (JTR) for multi-task learning with partially labeled samples.
        JTR ``stacks'' predictions and labels, encodes them into a single joint-task latent space, and minimizes the latent embedding distance. 
        JTR regularizes unlabeled task predictions using the labels of other tasks jointly in the latent space.
    }
    \label{fig:teaser}
    \vspace{-1em}
\end{figure}

In recent years, multi-task learning (MTL) has gained popularity in the field of machine learning. This approach involves training a model to perform multiple related tasks simultaneously, as opposed to traditional machine learning methods which train multiple independent models for each task. By leveraging commonalities across tasks, multi-task learning has the potential to improve overall performance while reducing inefficiencies and redundancies in the learning process. In particular, dense prediction tasks such as semantic segmentation
and depth estimation
have been particularly promising areas for multi-task learning, as these tasks typically require large models and significant computational resources.
Extensive research has been conducted in various directions to achieve effective multi-task learning~\cite{caruana1997multitask, ruder2017overview, li2022label2label, vandenhende2021multi}, including architecture design
\cite{misra2016cross, ruder2019latent, bragman2019stochastic, lu2017fully, guo2020learning, bruggemann2020automated, vandenhende2020branched}
and optimization strategies~\cite{kendall2018multi, gong2019comparison, xin2022current, chen2018gradnorm, yu2020gradient, navon2022multi}.
While these studies have demonstrated the effectiveness of multi-task learning, their applicability is limited by data availability. Training multi-task models requires high quality data, which is often expensive and time-consuming to acquire.

Collecting data for multi-task learning presents two primary challenges. The first challenge is the cost of annotating dense labels. 
For instance, annotating a segmentation mask for a single image in the Cityscapes dataset took over 1.5 hours on average~\cite{cordts2016cityscapes}. 
Single task semi-supervised learning presents a potential solution to this problem---however, despite extensive research on semi-supervised learning~\cite{berthelot2019mixmatch, mendel2020semi, olsson2021classmix, guizilini2020robust}, its application to multi-task learning remains underexplored.
Second, for tasks which involve predicting 
other modalities from images (such as depth or surface normal estimation), obtaining data introduces the challenge of sensor alignment and synchronization, requiring an expensive sensor fusion system. Building a sensor fusion system with high quality calibration and synchronization is difficult and necessitates specialized research.~\cite{yeong2021sensor}. One way to overcome this problem is to collect data for various subsets of tasks in a particular domain and train a model to perform all of the tasks jointly. However, multi-task learning without fully labeled data requires dealing with partial supervision, as each data sample is labeled for different tasks. Previous works have attempted this form of partial supervision for multi-task models in multiple ways: for instance, single-task semi-supervised learning methods regularize models through feature or output consistency with perturbed inputs for unlabeled tasks~\cite{jeong2019consistency, sohn2020fixmatch, ouali2020semi, fan2023revisiting,ghiasi2021multi}. While these approaches supplement the lack of supervision, they do not fully utilize the underlying cross-task relationships which are present in multi-task learning. To remedy this issue, recent pair-wise multi-task methods perform regularization across every pair of labeled and unlabeled tasks~\cite{zamir2018taskonomy, wang2019neural, standley2020tasks, zamir2020robust}. While such methods have shown promising results, their performance and poor scalability due to their quadratic complexity leaves much to be desired.

To this end, we propose a new approach for achieving data-efficient multi-task learning using partially labeled data. Our method, Joint-Task Regularization (JTR), encodes predictions and labels for multiple tasks into a single joint-task latent space. The encoded features are then regularized by a distance loss in the joint-task latent space, propagating gradients through all task branches at once. Our approach has two main advantages:
first, information flows across multiple tasks during regularization with backpropagation through a learned encoder, yielding a better metric space than naive spaces such as pixel-wise Euclidean or cosine distance.
Second, our method scales favorably relative to the numbers of tasks at hand. Unlike previous methods which regularize tasks in a pair-wise fashion~\cite{zamir2018taskonomy, wang2019neural, standley2020tasks, zamir2020robust}, JTR has linear complexity---this gives JTR a notable advantage in datasets with a large number of tasks over pair-wise task regularization methods which scale quadratically.

In our experiments, we demonstrate the effectiveness of our method on variations of three popular multi-task learning benchmarks, namely NYU-v2, Cityscapes, and Taskonomy.
We also conduct additional performance comparisons and ablation studies to inspect JTR's key characteristics.

To summarize, our main contributions are as follows: 

\begin{itemize}
    \item We investigate the underexplored problem of label-efficient multi-task learning with partially labeled data.
    \item We propose JTR, a model-agnostic technique which introduces a joint-task space to regularize all tasks at once.
    \item We extensively benchmark JTR on the NYU-v2, Cityscapes, and Taskonomy datasets to demonstrate its advantages over the current state-of-the-art. 
\end{itemize}

\section{Related Works}

\paragraph{Multi-Task Learning}
Multi-task learning aims to improve models' performance and generalization capabilities on individual tasks by exploiting commonalities and interdependencies between tasks through shared representations.
Several methods have been proposed to achieve effective multi-task learning through architectural modifications. 
For example, \cite{misra2016cross, ruder2019latent, bragman2019stochastic, chen2023mod, chen2023adamv} use multiple task experts and interconnect them to allow information and representation learning to flow across multiple tasks. 
Meanwhile,~\cite{lu2017fully, guo2020learning, bruggemann2020automated, vandenhende2020branched, zhang2022automtl, choi2023dynamic, aich2023efficient} adopt a strategy which gradually expands the depth of the model being trained, allowing the network to learn task-specific representations in a more resource efficient manner.
Various other techniques have been proposed for multi-task learning, including the attention mechanism~\cite{zhang2018joint, liu2019end, bruggemann2021exploring}, knowledge distillation~\cite{xu2018pad, vandenhende2020mti, ghiasi2021multi}, task pattern propagation~\cite{zhang2019pattern, zhou2020pattern}, generative models~\cite{bao2022generative}, and transformers~\cite{xu2022mtformer, ye2022inverted, fanm3vit}.
Another line of work tackles multi-task learning from an optimization perspective. %
Numerous optimization techniques have been proposed: some examples include loss weighting~\cite{kendall2018multi, gong2019comparison, xin2022current}, gradient normalization~\cite{chen2018gradnorm}, gradient dropout~\cite{chen2020just}, gradient surgery~\cite{yu2020gradient}, Nash Bargaining solutions~\cite{navon2022multi}, Pareto-optimal solutions~\cite{sener2018multi, lin2019pareto, momma2022multi}, gradient alignment~\cite{Senushkin_2023_CVPR}, and curriculum learning~\cite{guo2018dynamic, igarashi2022multi}.
Recently, with the rise of large-scale models and the pretrain-finetune paradigm, adapter-based multi-task finetuning methods~\cite{liangeffective, liu2022polyhistor} have also been introduced. 
Although extensive studies have been conducted on multi-task learning, these methods still require fully supervised training data which is difficult to obtain in practical scenarios due to high costs.

\vspace{-0.5em}
\paragraph{Partially Labeled Multi-Task Learning}
The advent of multi-task learning with partially labeled data has received limited attention in existing literature. Early works focused on multi-task learning with shallow models by employing parameter sharing~\cite{liu2007semi, zhang2009semi} or convex relaxation~\cite{wang2009semi}. More recently, several studies have tackled multi-task learning with missing labels based on deep models. For instance, Chen~\etal~\cite{chen2020multi} propose to use a consistency loss across complementary tasks such as shadow edges, regions, and counts to tackle a shadow detection problem by leveraging unlabeled samples. However, this method is limited to tasks which have explicit connections to each other. 
Others have attempted more general multi-task learning with partially labeled data with techniques such as domain discriminators~\cite{wang2022semi} and cross-task regularization by joint pair-wise task mappings~\cite{li2022learning}.
Furthermore, an orthogonal work by Borse \etal~\cite{borse2023dejavu} adds an auxiliary conditional regeneration objective to MTPSL~\cite{li2022learning} to improve performance.

Although these approaches address multi-task learning with partially labeled data, their performance leaves much to be desired. Additionally, existing methods are often challenging to implement in practice and scale poorly to larger datasets with many tasks. For example, MTPSL~\cite{li2022learning} incorporates FiLM~\cite{perez2018film} in an effort to offset its quadratic complexity, but this introduces additional hyperparameters, architecture-level modifications, and optimizers to the training process. In this work, we investigate a new avenue for achieving effective multi-task learning 
without compromising on simplicity and scalability.

\vspace{-1.0em}
\paragraph{Cross-Task Relations} Given that different visual perception tasks are often correlated with each other~\cite{zamir2018taskonomy,zamir2016generic}, there is growing interest in exploring cross-task relations~\cite{liu2010single,lu2021taskology,zamir2020robust,guizilini2020semantically,li2022learning}. For example, Taskonomy~\cite{zamir2018taskonomy} exposes relationships among different visual tasks and models the structure in a latent space. Cross-task relations have also been exploited for domain adaptation~\cite{patel2015visual,zamir2020robust,saha2021learning}, and cross-task consistency learning~\cite{zamir2020robust} attains better generalization to out-of-distribution inputs. Saha \etal~\cite{saha2021learning} propose a cross-task relation layer to encode task dependencies between the segmentation and depth predictions, and successfully improve model performance in an unsupervised domain adaptation setting. Some studies utilize cross-task relations to improve the performance of a single task~\cite{chen2019towards,guizilini2020semantically}. Guizilini \etal~\cite{guizilini2020semantically} leverage semantic segmentation networks for self-supervised monocular depth prediction. Cross-task consistency is also extensively investigated in MTL. Taskology~\cite{lu2021taskology} designs a consistency loss to enforce the logical and geometric structures of related tasks, reducing the need for labeled data. Most relevant to our work, Li \etal~\cite{li2022learning} leverage task relations by mapping each task pair to a joint pair-wise task space for multiple dense prediction tasks on partially annotated data. In this work, we propose a simple and scalable approach to perform regularization in a joint-task space by leveraging implicit cross-task relations.

\section{Method}

\label{method}

\subsection{Definitions}
Let $\mathcal{M}$ be the set of all task indices $\{ 1, \cdots, K \}$ where $K$ is the number of tasks. We define $\mathcal{D}$ as the set of $N$ training samples for $K$ tasks. For each training image $x$, let $\mathcal{T}_x$ and $\mathcal{U}_x$ be the sets containing indices of tasks which are labeled and unlabeled, respectively (such that $\mathcal{T}_x \cup \mathcal{U}_x = \mathcal{M}$). Let $x \in \mathbb{R}^{3 \times H \times W}$ denote a $H \times W$ RGB input image in $\mathcal{D}$, and let its corresponding dense label for task $t \in \mathcal{M}$ be denoted by $y^t \in \mathbb{R}^{O_t \times H \times W}$ where $O_t$ is the number of output channels for task $t$. As is standard in MTL literature~\cite{kendall2018multi, chen2018gradnorm, sener2018multi}, we seek to fit a common backbone feature extractor $f_{\phi}: \mathbb{R}^{3 \times H \times W} \rightarrow \mathbb{R}^{C \times H' \times W'}$ parameterized by $\phi$ where $C$, $H'$, $W'$ are the channel, height, and width of the extracted feature map, respectively (where typically $H' < H$ and $W' < W$). 
 We also seek to fit multiple task-specific decoders $h_{\psi^t}: \mathbb{R}^{C \times H' \times W'} \rightarrow \mathbb{R}^{O_t \times H \times W}$, each separately parameterized by task-specific weights $\psi^t$. For each target task $t \in \mathcal{M}$, the model's prediction can be expressed as $\hat{y}^t(x) = h_{\psi^t} \circ f_{\phi}(x)$. We also denote any variable with detached gradients using  $\langle$angular brackets$\rangle$.

\subsection{Baselines for MTL with Partially Labeled Data}

\paragraph{Supervised Multi-Task Learning}
A na\"ive way to fit $\hat{y}^t$ for all tasks is to optimize parameters $\phi$ and $\psi^t$ for all labeled tasks $t \in \mathcal{T}_{x_n}$ for all items $x_n$ in the training dataset $\mathcal{D}$ as follows: 
\begin{equation}
    \min_{\phi, \psi} 
    \frac{1}{|\mathcal{D}|} \displaystyle\sum\limits_{(x,y) \in \mathcal{D}} \displaystyle\sum\limits_{t \in \mathcal{T}_x} \frac{1}{|\mathcal{T}_x|}\mathcal{L}^t(\hat{y}^t(x), y^t)
    \label{eq:naive}
\end{equation}
where $\mathcal{L}^t$ is a differentiable loss function for a specific task $t$. Here, weights for the backbone feature extractor $\phi$ are learned using all labeled images while task-specific weights $\psi^t$ for each task $t \in \mathcal{M}$ are only learned for dataset examples $x$ which are labeled for task $t$ ($t \in \mathcal{T}_{x}$).

\begin{figure*}[t]
    \centering
    \includegraphics[width=1.0\linewidth]{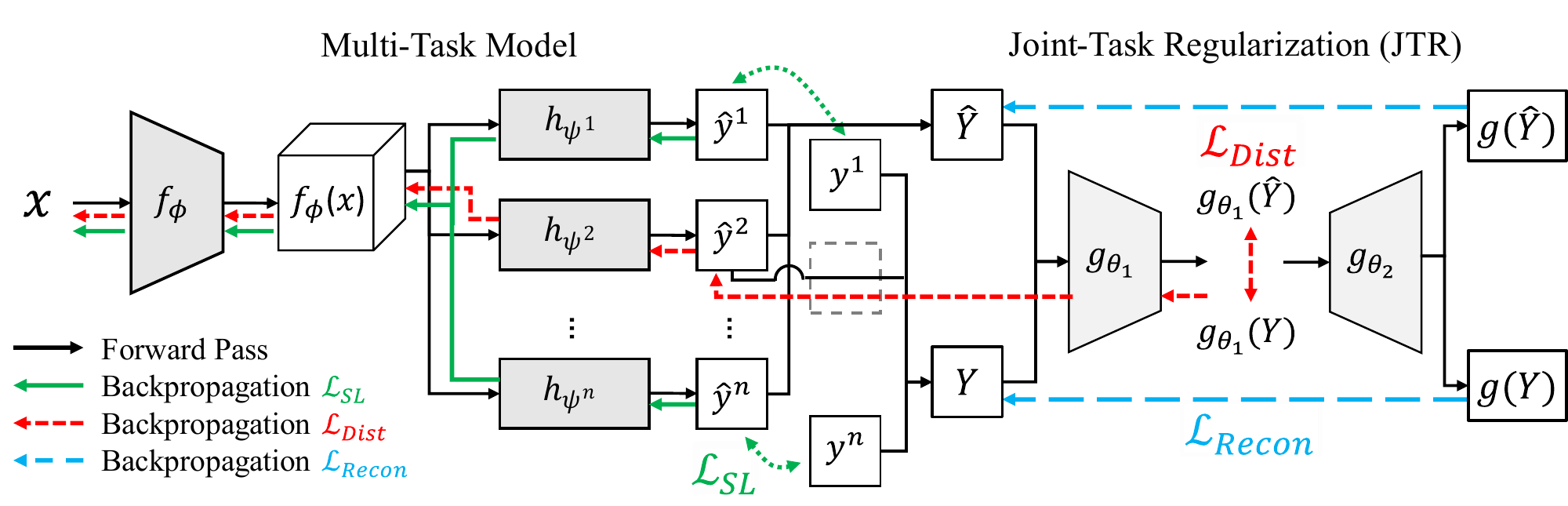}
    \vspace{-2em}
    \caption{Overview of JTR. We define an input $x$, a shared feature extractor $f_{\phi}$, task-specific decoders $h_{\psi^t}$, and output predictions $\hat{y}^t$. 
    For labeled tasks, supervised loss ($\mathcal{L}_{SL}$) is applied. 
    Then, predictions ($\hat{y}^t$) are concatenated to form $\hat{Y}$, while labels ($y^t$) are concatenated to form a target tensor $Y$ (using $\hat{y}^t$ as a substitute is no label exists). The JTR encoder $g_{\theta_1}$ encodes predictions from multiple tasks into one joint-task latent space. $\mathcal{L}_{Dist}$ enforces $\hat{Y}$'s latent embedding to be close to that of $Y$. Gradients from $\mathcal{L}_{Dist}$ apply to $g_{\theta_1}$ and $h_{\psi^t}$. Gradients from $\mathcal{L}_{Recon}$ apply to $g_{\theta_1}$ and $g_{\theta_2}$, preventing $g_{\theta_1}$ from learning a trivial solution (\ie encoding all inputs to a single point).
    }
    \label{fig:overview}
\end{figure*}

\paragraph{SSL with Consistency Regularization}
\label{sec:ssl-conreg}
One common strategy to improve upon na\"ive supervised learning is to apply Semi-Supervised Learning (SSL). In particular, Consistency Regularization (CR) is a common method of leveraging data augmentation to encourage models to output consistent predictions across multiple different augmentations of unlabeled data~\cite{jeong2019consistency, sohn2020fixmatch, ouali2020semi, fan2023revisiting}. More concretely, Eq. \ref{eq:naive} is modified as follows:
\begin{equation}
    \min_{\phi, \psi} 
    \frac{1}{|\mathcal{D}|} \displaystyle\sum\limits_{(x,y) \in \mathcal{D}} \left(\begin{matrix}
        {\displaystyle\sum\limits_{t \in \mathcal{T}_{x}} \frac{1}{|\mathcal{T}_{x}|} \mathcal{L}^t(\hat{y}^t(x), y^t)  }\\
        +\displaystyle\sum_{t\in \mathcal{U}_{x}}\frac{1}{|\mathcal{U}_{x}|}\mathcal{L}_{u}(\hat{y}^{t}(\mathcal{A}(x)),\hat{y}^{t}(\mathcal{A}(x)))
    \end{matrix}\right)
    \label{eq:naive-ssl}
\end{equation}
where $\mathcal{L}_u$ is an unsupervised loss function (\eg L2 loss) and $\mathcal{A}_r: \mathbb{R}^{3 \times H \times W} \rightarrow \mathbb{R}^{3 \times H \times W}$ is a function which applies augmentation to an image $x \in \mathbb{R}^{3 \times H \times W}$. 
We recognize that there are more advanced SSL methods which exploit task-specific properties such as SSL for semantic segmentation~\cite{mendel2020semi, olsson2021classmix} and depth estimation~\cite{guizilini2020robust} in single-task learning settings. 
However, it is commonly understood that applying pseudo-labeling methods in multi-task settings is a major challenge because most semi-supervised loss functions are designed based on task-specific properties, and it is unclear how to effectively combine them when working with multiple tasks simultaneously~\cite{li2022learning}. 
Therefore, following previous works, we primarily use Consistency Regularization as the baseline SSL method throughout this work (for completeness, we also benchmark SSL with pseudo-labels in \Sref{sec:ablation-baseline}).

\paragraph{Pair-Wise Task Relations}

While Eq. \ref{eq:naive-ssl} indeed incorporates partially labeled data for training, it does not leverage implicit relationships and commonalities between tasks. To exploit such potentially useful task relationships, some previous works have proposed to learn pair-wise mappings or construct explicit graph connections between pairs of tasks~\cite{zamir2018taskonomy, wang2019neural, standley2020tasks, zamir2020robust}. Most relevant to our work is MTPSL~\cite{li2022learning} which creates a ``mapping space'' for each pair of tasks. Using these task pair-wise mapping spaces, MTPSL seeks to maximize the similarity of an unlabeled task prediction ``mapped'' to a labeled task as follows:
\begin{equation}
    \begin{array}{cc}
        \displaystyle\frac{1}{|\mathcal{D}|} \displaystyle\sum\limits_{(x,y) \in \mathcal{D}}
        \frac{1}{|\mathcal{U}_{x}|}\sum_{s\in \mathcal{U}_{n}, t \in \mathcal{T}_{n}}\mathcal{L}_{ct}\left(
            \begin{array}{cc}
                m_{\vartheta_{s}^{st}}(\hat{y}^{t}(x_{n})),\\
                m_{\vartheta_{t}^{st}}(y^t)
            \end{array}
        \right)
    \end{array}
\end{equation}
where $\mathcal{L}_{ct}$ is the cosine distance function ($\mathcal{L}_{CT}(\mathbf{a}, \mathbf{b}) = 1 - \frac{\mathbf{a} \cdot \mathbf{b}}{|\mathbf{a}||\mathbf{b}|}$), $m$ is a learned mapping function (typically an encoder similar to $g_{\theta_1}$), and each $\vartheta_{t}^{st}$ is a set of parameters which parameterizes $m$. MTPSL additionally incorporates FiLM~\cite{perez2018film} to reduce number of parameters required for parameterizing $\vartheta$ for each task pair $(s, t)$. However, because $\mathcal{L}_{CT}$ is still computed for task pairs, MTPSL suffers from quadratic complexity and is unable to leverage information which may arise from relationships among more than $2$ tasks.

\subsection{Joint-Task Regularization (JTR)}

To remedy the aforementioned issues with pair-wise mapping schemes like MTPSL, JTR completely avoids modeling task relations in pairs by instead learning a regularization space as a single function of all $K$ tasks.
More concretely, for each input image $x$ and corresponding label $y$, we generate a noisy prediction tensor $\hat{Y}_x$ by stacking predictions ($\hat{y}^t$) for all tasks along the channel dimension. 
Then, we create a reliable target tensor $Y_x$ by stacking labels ($y^t$) if task $t$ is labeled, using predictions ($\hat{y}^t$) as a substitute when no label exists for task $t$. 
Formally, let $\hat{Y}_x$ and $Y_{x}$ be defined as follows:
\begin{equation}
        \hat{Y}_{x} = 
            [
                \hat{y}^{1}_x
                \;
                ;
                \;
                \hdots
                \;
                ;
                \;
                \hat{y}^{K}_x
            ],
            \;\;
        Y_{x} = 
            [
                \delta(x, 1)
                \;
                ;
                \;
                \hdots
                \;
                ;
                \;
                \delta(x, K)
            ]
\end{equation}
where 
$\delta(x, t) = \begin{cases} 
y^t_x & t \in \mathcal{T}_{x} \\
\hat{y}^t_x & t \notin \mathcal{T}_{x}
\end{cases}$. The shapes of $\hat{Y}_{x}$ and $Y_{x}$ are $\left(\sum_{t=1}^K O_t\right) \times H \times W$, where $O_t$ is the number of channels for task $t$.
Then, we define the JTR encoder-decoder as $g: \mathbb{R}^{C \times H \times W} \rightarrow \mathbb{R}^{C \times H \times W} = (g_{\theta_2} \circ g_{\theta_1})$. Here, $g_{\theta_1}: \mathbb{R}^{C \times H \times W} \rightarrow \mathbb{R}^{C' \times H' \times W'}$ and $g_{\theta_2}: \mathbb{R}^{C' \times H' \times W'}\rightarrow \mathbb{R}^{C \times H \times W}$ for some $C'$. We call the output space of $g_{\theta_1}$ the ``joint-task space,'' since it represents information about all tasks at once. In this setup, $g$ forms an auto-encoder architecture which encodes and reconstructs $Y_x$ and $\hat{Y}_x$ across the joint-task space as the bottleneck, preventing the encoder $g_{\theta_1}$ from learning a trivial joint-task space. Now, we can define a loss term to minimize the distance between $Y_x$ and $\hat{Y}_x$ in the joint-task space while also fitting $g_{\theta_1}$ and $g_{\theta_2}$:
\begin{equation}
    \mathcal{L}_{JTR} = \frac{1}{|\mathcal{D}|} \displaystyle\sum\limits_{(x,y) \in \mathcal{D}}  \left(
    \begin{array}{l}
        \mathcal{L}_{Dist}(g_{{\theta_1}}(\hat{Y}_{x}), g_{\theta_1}(Y_x )) \\
        +\mathcal{L}_{Recon}(g({\langle\hat{Y}_x}\rangle),{\langle\hat{Y}_x}\rangle) \\
        +\mathcal{L}_{Recon}(g(\langle{Y}_{x}\rangle), \langle{Y}_{x}\rangle)\\
    \end{array}
    \right)
    \label{eq:l_ae}
\end{equation}
Here, the joint latent distance loss $\mathcal{L}_{Dist}$ is some distrance metric (\eg the cosine distance function $\mathcal{L}_{Dist}(\mathbf{a}, \mathbf{b}) = 1 - \frac{\mathbf{a} \cdot \mathbf{b}}{|\mathbf{a}||\mathbf{b}|}$), and $\mathcal{L}_{Recon}$ measures the quality of the decoder's reconstruction using the task-specific loss functions $\mathcal{L}^t$ for all tasks $t \in \mathcal{M}$. We also $\langle$detach$\rangle$ the $\mathcal{L}_{Recon}$ terms' gradients to prevent the model from outputting degenerate predictions to trivialize the reconstruction task. An illustration of these components of JTR is provided in \Fref{fig:overview}.

Now, we can sum the supervised loss and the JTR loss term from Eq. \ref{eq:l_ae} to get the following optimization objective:
\begin{equation}
    \min_{\phi, \psi, \theta_1, \theta_2} \left(\mathcal{L}_{SL} + \mathcal{L}_{JTR}\right)
    \label{eq:xtae}
\end{equation}
where $\mathcal{L}_{SL}$ is the supervised loss in Eq. \ref{eq:naive}.

Since $Y_x$ and $\hat{Y}_x$ aggregate predictions and labels for all tasks into a single representation, JTR's distance loss term in the joint-task space propagates gradients to all tasks jointly. This has both performance and efficiency benefits: for one, JTR is able to leverage relationships which may arise from non-obvious combinations of tasks, giving it an advantage over existing methods which only model inter-task dependencies by pairs~\cite{zamir2018taskonomy, wang2019neural, standley2020tasks, zamir2020robust}. Second, JTR achieves linear complexity with respect to the number of tasks, differentiating itself from existing methods which scale quadratically. As our experiments show in the following section, these benefits allow JTR to perform significantly more favorably than existing methods in a variety of partially labeled multi-task learning settings.

\section{Experiments}

\subsection{Experimental Setup}

    To benchmark JTR against existing partially labeled MTL methods, we conduct experiments on three common multi-task dense prediction benchmarks: namely NYU-v2 \cite{silberman2012indoor, liu2019end}, Cityscapes \cite{cordts2016cityscapes}, and Taskonomy~\cite{zamir2018taskonomy}. These public research datasets are intended for benchmarking purposes, so each data sample is fully labeled for all tasks. To simulate more realistic partially labeled scenarios, we follow \cite{li2022learning} in synthetically generating ``randomlabels'' and ``onelabel'' settings for all three datasets. Throughout our work, ``randomlabels'' means each dataset example has a random number of labels between $1$ and at most $K-1$ (where $K$ is the number of tasks), while ``onelabel'' means each example has exactly one label for a randomly chosen task.
    For the Taskonomy benchmark, we follow previous works~\cite{standley2020tasks} in training multi-task models on seven tasks: semantic segmentation, depth, surface normals, edge occlusions, reshading, 2D keypoints, and edge textures. 
    Additionally, we introduce one more partially labeled setting named ``halflabels'' for the Taskonomy experiment. In ``halflabels,'' each dataset sample has $3$ or $4$ labels out of $7$ tasks. This is a more balanced partially labeled setting than the ``randomlabels'' setting, while the mean number of labels in both settings is the same. The Taskonomy experiments are conducted using the `Tiny' dataset split.

    We base the implementation of JTR (as well as all baselines) on the public code repository provided by Li \etal~\cite{li2022learning,mtpsl_gh}. Across all experiments, 
    we use SegNet \cite{badrinarayanan2017segnet} as the MTL backbone. For JTR, we use the same SegNet architecture as the MTL backbone with only the input and output channel dimensions modified as needed. We also use the SegNet architecture for the ``Direct Cross-Task Mapping'' and ``Perceptual Cross-Task Mapping'' baselines as formulated by Li \etal~\cite{li2022learning} and Zamir \etal~\cite{zamir2020robust}. We use cross-entropy loss for semantic segmentation, L1 norm loss for depth estimation, and cosine similarity loss for surface normal estimation. 
    For the four additional tasks in the Taskonomy experiment, we use L1-norm loss following Standley \etal~\cite{standley2020tasks}.
    Additionally, we apply RandAugment~\cite{cubuk2020randaugment} across our experiments. We use mostly identical hyper-parameters including the learning rate and the optimizer as the code for MTPSL~\cite{li2022learning,mtpsl_gh} (see Supplementary Material for more details). For evaluation metrics, we use mean intersection over union (mIoU), absolute error (aErr), and mean angle error (mErr) for semantic segmentation, depth estimation, and surface normal estimation tasks, respectively. 
    For the four additional tasks in Taskonomy, we use absolute error (aErr).
    We also report the mean percentage ``improvement'' (relative performance increase or relative error decrease) of each benchmarked method over the corresponding multi-task Supervised MTL baseline (averaged across all tasks) as ``$\overbar{\Delta\%}$.''

\begin{table}
    \centering
    \tablescale{{
		\begin{tabular}{lccccccccccc}
            \toprule
            Method & Seg. $\uparrow$ & Depth $\downarrow$ & Norm. $\downarrow$ & $\overbar{\Delta\%}$ $\uparrow$ \\
            \midrule
            \multicolumn{5}{c}{\textit{fully labeled}}\\
            Supervised MTL* & 38.91 & 0.5351 & 28.57 & --- \\
            \midrule
            \multicolumn{5}{c}{\textit{onelabel}}\\
            Supervised MTL* & 25.75 & 0.6511 & 33.73 & +0.000 \\
            Consistency Reg.* & 27.52 & 0.6499 & 33.58 & +2.501 \\
            Direct Map*~\cite{zamir2020robust} & 19.98 & 0.6960 & 37.56 & --13.55  \\
            Perceptual Map*~\cite{zamir2020robust} & 26.94 & 0.6342 & 34.30 & +1.842 \\
            MTPSL \cite{li2022learning}& 30.40 & 0.5926 & 31.68 & +11.04 \\
            \textbf{Ours (JTR)}& \textbf{31.96} & \textbf{0.5919} & \textbf{30.80} & \textbf{+13.97} \\
            \multicolumn{5}{c}{\textit{depth-extended ``onelabel''}}\\
            Supervised MTL & 23.43 & 0.6918 & 39.44 & --10.73 \\
            Consistency Reg. & 23.94 & 0.6840 & 40.12 & --10.34 \\
            MTPSL \cite{li2022learning} & \textbf{33.80} & \textbf{0.4794} & 37.50 & +15.49 \\
            \textbf{Ours (JTR)} & 33.63 & 0.4898 & \textbf{31.37} & \textbf{+20.79} \\
            \midrule
            \multicolumn{5}{c}{\textit{randomlabels}}\\
            Supervised MTL* & 27.05 & 0.6624 & 33.58 & +0.000 \\
            Consistency Reg.* & 29.50 & 0.6224 & 33.31 & +5.300 \\
            Direct Map*~\cite{zamir2020robust} & 29.17 & 0.6128 & 33.63 & +5.060 \\
            Perceptual Map*~\cite{zamir2020robust} & 32.20 & 0.6037 & 32.07 & +10.80 \\
            MTPSL \cite{li2022learning}& 35.60 & 0.5576 & 29.70 & +19.66 \\
            \textbf{Ours (JTR)} & {\bf 37.08} & {\bf 0.5541} & {\bf 29.44} & \textbf{+21.92} \\
            \bottomrule
		\end{tabular}
    }}
    \caption{Partially labeled MTL results on NYU-v2. Results marked * are directly quoted from \cite{li2022learning} (see \Sref{sec:explaining-results} for details).}
    \label{tab:nyuv2}
\end{table}
%
%
%
%
%
%
%
%
%
%
%
%
%
%
%
%
%
%
%
%
%
%
%
%
%
%
%
%
%
%
%
%
%
%
%
%
%
%
%
%
%
%
%
%
%
%
%
%
%
%
%
%
%
%
%
%
%
%
%
%
%
%
%
%
%
%
%
%
%

\begin{table}[t]
    \centering
    \tablescale{
    {
		\begin{tabular}{llccccccccc}
            \toprule
            Scenario & Method & Seg. $\uparrow$ & Depth $\downarrow$ & $\overbar{\Delta\%}$ $\uparrow$ \\
            \midrule
            \multirow{1}{*}{\textit{fully labeled}} & 
            Supervised MTL* & 72.70 & 0.0163 & --- \\
            \midrule
            \multirow{4}{*}{\textit{onelabel}} & Supervised MTL* & 69.50 & 0.0186 & +0.000 \\
            & Consistency Reg.* & 71.67 & 0.0178 & +3.712 \\
            & MTPSL \cite{li2022learning} & 72.09 & 0.0168 & +6.702 \\
            & \textbf{Ours (JTR)} & \textbf{72.33} & {\bf 0.0163} & \textbf{+8.219} \\
            \bottomrule
		\end{tabular}
    }
    }
    \caption{Partially labeled MTL results on Cityscapes. Results marked * are directly quoted from \cite{li2022learning} (see \Sref{sec:explaining-results} for details).
    }
    \label{tab:cityscapes}
    \vspace{-1em}
\end{table}

\begin{table*}
    \centering
    \tablescale{{
		\begin{tabular}{llccccccc a}%
            \toprule
            Scenario & Method & $\Delta$Seg.$\%$  & $\Delta$Depth$\%$ & $\Delta$Norm.$\%$ &
            $\Delta$Occl.$\%$ & $\Delta$Resh.$\%$ & $\Delta$KeyP.$\%$ & $\Delta$Text.$\%$ &
            $\overbar{\Delta\%}$ $\uparrow$
            \\
            \midrule
            \multirow{3}{*}{\textit{onelabel}}& Consistency Reg.            
            & +0.080 & +6.106 & +4.957 & +1.124 & +3.905 & --8.723 & +17.63 & +1.567 \\
            & MTPSL \cite{li2022learning} 
            & +0.646 & +4.172 & +6.201 & +0.829 & +7.540 & --11.75 & +23.94 & +1.973 \\ 
            & \textbf{Ours (JTR)}         
            & +0.381 & +8.477 & +6.236 & +1.076 & +6.223 & --1.222 & +35.03 & \textbf{+3.512} \\
            \midrule
            \multirow{3}{*}{\textit{randomlabels}}& Consistency Reg.
            & --0.058 & +8.438 & +1.218 & +0.531 & +4.460 & +6.622 & +14.86 & +2.254 \\
            & MTPSL \cite{li2022learning} 
            & +0.311 & +17.70 & +2.355 & +1.349 & +10.08 & --20.68 & +9.771 & +1.261 \\
            & \textbf{Ours (JTR)}         
            & +0.193 & +12.14 & +9.061 & +2.030 & +9.740 & +1.955 & +30.40 & \textbf{+4.095}\\
            \midrule
            \multirow{3}{*}{\textit{halflabels}} & Consistency Reg.            
            & +0.578 & +4.173 & +5.884 & +1.252 & +4.493 & +11.43 & +9.415 & +2.326 \\
            & MTPSL \cite{li2022learning} 
            & +0.656 & +7.723 & +10.49 & +2.496 & +9.100 & +0.552 & +19.55 & +3.160 \\
            & \textbf{Ours (JTR)}         
            & +0.776 & +11.05 & +13.28 & +3.055 & +10.98 & +9.442 & +32.62 & {\bf+5.075}  \\ 
            \bottomrule
		\end{tabular}
    }}
    \caption{Partially labeled MTL results on Taskonomy. Scores are relative percentage improvements over the ``Supervised MTL'' baseline.}
    \label{tab:taskonomy}
    \vspace{-0.5em}
\end{table*}
\subsection{Partially Labeled MTL}
\label{results:nyu}

\paragraph{NYU-v2}
We present our results with NYU-v2 under the ``onelabel'' and ``randomlabels'' settings in Tab. \ref{tab:nyuv2}. We observe that in both scenarios, JTR convincingly outperforms existing methods across all three tasks. Note that results marked * are quoted from \cite{li2022learning} to maintain consistency with results from previous works (for more details, please see our explanation of ``Copied Baselines'' in \Sref{sec:explaining-results}). 

\paragraph{Cityscapes}

We evaluate our method on Cityscapes under the ``onelabel'' setting in Tab. \ref{tab:cityscapes}. Similarly to our results with NYU-v2, JTR outperforms baseline methods across both segmentation and depth estimation tasks. One point to note is that the differences in performance between methods on Cityscapes is less prominent than on NYU-v2, suggesting that Cityscapes is a much less challenging dataset for partially labeled MTL. 
Nevertheless, JTR comfortably outperforms existing methods on all Cityscapes tasks.

%

%
%
%
%
%
%
%
%
%
%
%
%
            
%
%
%
%
%
%
%
%
%
%
%
%
%
%

\paragraph{Taskonomy}

We present our results using Taskonomy in three partially labeled settings in~\Tref{tab:taskonomy}. In this dataset, consistency regularization, MTPSL and JTR improve upon na\"ive Supervised MTL. 
Notably, the consistency regularization baseline and MTPSL enhance overall performance but compromise certain tasks to improve others. For example, the keypoint detection task suffers from large negative effects in the `onelabel' scenario.
In contrast, JTR enhance overall performance without large detriments to individual task performance metrics.
Our experiments indicate that some form of cross-task regularization is essential to prevent models from biasing for or against specific tasks.

JTR's margins of improvement on Taskonomy, while noticeable, are not as significant as with NYU-v2 and Cityscapes. This could be attributed to the larger scale of the Taskonomy dataset, which enhances the performance of the supervised baseline. 
Nonetheless, JTR is more effective in improving the overall MTL performance under different partially labeled scenarios than the competing methods.
Additionally, our method holds a significant advantage in terms of computational resource efficiency (further explained in~\Sref{sec:additional}).

\subsection{Depth-Extended NYU-v2}

In addition to the $795$ fully labeled images in multi-task NYU-v2, there are an additional $47584$ samples in the original NYU v2 dataset which are only labeled for the depth estimation task~\cite{silberman2012indoor, icra_2019_fastdepth}. This is dataset is approximately $60$ times larger than the training split of multi-task NYU-v2. 
To leverage this additional data, we combine the training splits of two datasets to curate a new split named ``depth-extended multi-task NYU-v2.'' 
In the context of our experimental setup, this is equivalent to the ``onelabel'' scenario---therefore, we can leverage this extra data by simply extending the ``onelabel'' split. This setup puts each method's resilience to task imbalance to the test.

Our results are presented in \Tref{tab:nyuv2} alongside the NYU-v2 ``onelabel'' results for direct comparison. We can notice that the inclusion of extra depth data has a negative effect on Supervised MTL and Consistency Reg. across all tasks (including depth estimation), highlighting the brittleness of na\"ive training to task imbalances. For MTPSL and JTR, both outperform their standard ``onelabel'' counterparts in the segmentation and depth tasks, with MTPSL having a slight edge. However, MTPSL performs very poorly in the surface normal estimation task while JTR maintains comparable performance, yielding a significantly better overall $\overbar{\Delta\%}$ score. These results show that JTR is highly resilient in scenarios with task imbalances (even without additional task-weighting tricks), underscoring the stability and practicality of our approach.

\begin{table}
    \centering
    \tablescale{{
		\begin{tabular}{lccccccccccc}
            \toprule
            Method & Seg. $\uparrow$ & Depth $\downarrow$ & Norm. $\downarrow$ & $\overbar{\Delta\%}$ $\uparrow$ \\
            \midrule
            \multicolumn{5}{c}{\textit{70\% randomlabels + 30\% unlabeled}}\\
            Supervised MTL & 26.36 & 0.6574 & 33.02 & +0.000\\
            Consistency Reg. & 29.89 & 0.6136 & 31.93 & +7.785 \\
            MTPSL~\cite{li2022learning} &  31.92& 0.6071 & 31.59  & +11.03 \\
            \textbf{Ours (JTR)} & \textbf{32.20} & \textbf{0.5912} & \textbf{31.01}  & \textbf{+12.77} \\
            \midrule
            \multicolumn{5}{c}{\textit{50\% randomlabels + 50\% unlabeled}}\\
            Supervised MTL & 23.68 & 0.6988 & 35.00  & +0.000\\
            Consistency Reg. & 25.16 & 0.6543 & 32.54  & +6.549 \\
            MTPSL~\cite{li2022learning} & 28.18 & 0.6393 & 33.04  & +11.04 \\
            \textbf{Ours (JTR)} & {\bf 28.69} & \textbf{0.6268} & \textbf{31.52} & \textbf{+13.80} \\
            \midrule

            \multicolumn{5}{c}{\textit{30\% randomlabels + 70\% unlabeled}}\\
            Supervised MTL & 18.57 & 0.7288 & 37.49  & +0.000\\
            Consistency Reg. & 20.60 & 0.7135 & 34.81  & +6.727 \\
            MTPSL~\cite{li2022learning} & 23.33 & 0.6952 & 34.81 & +12.46 \\
            \textbf{Ours (JTR)} & \textbf{24.05} & \textbf{0.6945} & \textbf{34.25} &  \textbf{+14.29}\\
            \bottomrule
		\end{tabular}
    }}
    \caption{Partially labeled multi-task learning results on NYU-v2. Results marked * are directly quoted from \cite{li2022learning}.}
    \label{tab:nyuv2-ssl}
\end{table}

\begin{table}[t]
    \centering
    \tablescale{
    {
		\begin{tabular}{llccccccccc}
            \toprule
            Scenario & Method & Seg. $\uparrow$ & Depth $\downarrow$ & $\overbar{\Delta\%}$ $\uparrow$ \\
            \midrule
            \multirow{4}{*}{\textit{\makecell[l]{70\% onelabel,\\30\% unlabeled}}} &  Supervised MTL           & 69.62 & 0.0180 &  +0.000\\
            & Consistency Reg.     & 69.65 & 0.0171 & +2.522 \\
            & MTPSL~\cite{li2022learning}                         & \textbf{71.48} & 0.0169 &  +4.391\\
            & \textbf{Ours (JTR)}           & 71.07 & \textbf{0.0161} & \textbf{+6.319} \\
            \midrule
            \multirow{4}{*}{\textit{\makecell[l]{50\% onelabel,\\50\% unlabeled}}} & Supervised MTL           & 67.56 & 0.0196 & +0.000\\
            & Consistency Reg.     & 69.23 & 0.0174 & +6.848 \\
            & MTPSL~\cite{li2022learning}                         & \textbf{70.30} & 0.0174 & +7.640 \\
            & \textbf{Ours (JTR)}           & 69.76 & \textbf{0.0167} & \textbf{+9.026} \\
            \midrule
            \multirow{4}{*}{\textit{\makecell[l]{30\% onelabel,\\70\% unlabeled}}} & Supervised MTL           & 64.65 & 0.0217 & +0.000 \\
            & Consistency Reg.     & 66.43 & 0.0187 & +8.289  \\
            & MTPSL~\cite{li2022learning} & \textbf{68.09} & 0.0187 & +9.573 \\
            & \textbf{Ours (JTR)}        & 67.75 & \textbf{0.0179} & \textbf{+11.15} \\
            \bottomrule
		\end{tabular}
    }
    }
    \caption{Partially labeled MTL results on Cityscapes. Results marked * are directly quoted from \cite{li2022learning}.
    \vspace{-1.0em}
    }
    \label{tab:cityscapes-ssl}
\end{table}

\subsection{Partially (Un)labeled MTL}

To further test JTR's capabilities under diverse partial label settings, we generate variants of NYU-v2 and Cityscapes in which a portion of the dataset is randomly chosen to be completely unlabeled, and the remainder is labeled following the aforementioned ``randomlabels'' or ``onelabel'' configurations. To enable effective learning learning in these settings, we simply apply consistency regularization in the joint-task latent space when no labels are available. We implement this by making a small modification to Eq. \ref{eq:l_ae} for completely unlabeled samples (i.e. when $|\mathcal{T}_x| = 0$):
$$
\mathcal{L}_{Dist}(g_{{\theta_1}}(\hat{Y}_{x}), g_{\theta_1}(Y_x )) \rightarrow \mathcal{L}_{Dist}(g_{{\theta_1}}(\hat{Y}_{\mathcal{A}(x)}), g_{\theta_1}(Y_{\mathcal{A}(x)} ))
$$

For both datasets, we created 30\% unlabeled, 50\% unlabeled, and 70\% unlabeled data splits. The labeled portion of the data is kept consistent with the ``randomlabels'' and ``onelabel'' scenarios from NYU-v2 ``randomlabels'' and Cityscapes ``onelabel,'' respectively. All hyperparameters are kept constant across all unlabeled ratios.

\paragraph{NYU-v2}

Results are shown in \Tref{tab:nyuv2-ssl}. In these experiments, all three benchmarked methods improve upon na\"ive Supervised MTL across all label scenarios. 
However, JTR consistently outperforms baseline methods across all tasks. Some baselines nearly match JTR in specific tasks---for example, while the difference in depth aErr between MTPSL and JTR in the 70\% unlabeled experiment is marginal, JTR handily outperforms MTPSL in the segmentation and normal estimation tasks. Thanks to this ability to better balance multiple tasks, JTR obtains the best overall $\overbar{\Delta\%}$ score. One notable trend is that the $\overbar{\Delta\%}$ score improvement of JTR becomes more pronounced as the level of supervision decreases. This indicates that JTR is effectively taking advantage of its access to unlabeled data to maximize learning.

\vspace{-0.5em}
\paragraph{Cityscapes}

Results are shown in \Tref{tab:cityscapes-ssl}. We observe a similar trend with Cityscapes---JTR nearly matches MTPSL's segmentation mIoU while vastly surpassing all existing methods in depth estimation, and it achieves the highest overall $\overbar{\Delta\%}$ score across every label scenario. While satisfactory, we believe our results in Cityscapes can be further improved by adjusting several parameters to suit the reduction in available data. However, we refrain from varying any hyperparameters as the results already sufficiently demonstrate the efficacy of our approach.

\subsection{Additional Results}
\label{sec:additional}

\begin{table}[t]
    \centering
    \tablescale{
    {
		\begin{tabular}{lrrrr}
            \toprule
            \multirow{2}{*}{Method} & \multicolumn{2}{c}{SegNet} & \multicolumn{2}{c}{ResNet-50} \\
            & Time $\downarrow$ & VRAM $\downarrow$ & Time $\downarrow$ & VRAM $\downarrow$\\
            \midrule
            MTPSL \cite{li2022learning}& 8h\:20m & 17.5GiB & 17h\:40m & 21.5GiB\\
            \textbf{Ours (JTR)} & 9h\:00m & 17.5GiB & 10h\:50m & 16.2GiB \\
            \bottomrule
		\end{tabular}
    }
    }
    \caption{
        Time and VRAM requirements for training on NYU-v2 using a fixed batch size of $4$ on an NVIDIA A100 for $300$ epochs. See Supplementary Material for more details and results for the ResNet-50 experiment.
    }
    \label{tab:nyuperf}
\end{table}

\begin{table}[t]
    \centering
    \tablescale{
    {
		\begin{tabular}{lrrrr}
            \toprule
            \multirow{2}{*}{Method} & \multicolumn{2}{c}{Cityscapes} & \multicolumn{2}{c}{Taskonomy} \\
            & Time $\downarrow$ & VRAM $\downarrow$ & Time $\downarrow$ & VRAM $\downarrow$\\
            \midrule
            MTPSL \cite{li2022learning}& 22h\:10m & 14.4GiB & 223h\:10m  &  34.4GiB \\
            \textbf{Ours (JTR)} & 23h\:45m & 19.2GiB  &  105h\:00m  &  23.9GiB \\
            \bottomrule
		\end{tabular}
    }
    }
    \caption{
        Time and VRAM requirements for training on Cityscapes and Taskonomy using SegNet on an NVIDIA A100, with $300$ and $20$ epochs respectively and batch sizes $16$ and $8$ respectively.
    }
    \vspace{-1em}
    \label{tab:generalperf}
\end{table}

\paragraph{Computational Costs}

The computational cost of JTR is dependent on number of tasks and the size of the encoder/decoder used to learn the joint-task latent space. For small-scale datasets such as NYU-v2 ($3$ tasks) and Cityscapes ($2$ tasks), JTR consumes the same or more resources than MTPSL (\Tref{tab:nyuperf}, \Tref{tab:generalperf}). However, this is due to the ``upfront'' cost of having the extra encoder/decoder for learning the joint-task latent space. When scaling up to realistic scenarios like Taskonomy ($7$ tasks), JTR gains a monumental advantage over MTPSL in total training time 
and VRAM usage (\Tref{tab:generalperf}). Furthermore, JTR uses less resources than MTPSL when using more typical model architectures like ResNet-50 rather than SegNet (\Tref{tab:nyuperf}). This gap can be explained by MTPSL's use of layer-wise FiLM transformations~\cite{perez2018film}, which requires more VRAM for more complex network architectures. We further elaborate on these results in the Supplementary Material.

\begin{table}
    \centering
    \tablescale{
        \begin{tabular}{lcccc}
            \toprule
            Method & Seg. $\uparrow$ & Depth $\downarrow$ & Norm. $\downarrow$ & $\overbar{\Delta\%}$ $\uparrow$\\
            \midrule
            \multicolumn{5}{c}{\textit{NYU-v2 ``randomlabels''}}\\
            Supervised MTL & 27.05 & 0.6624 & 33.58 & +0.000\\
            MTPSL w/o reg. \cite{li2022learning} & 33.51 & 0.5767 & 34.00 & +11.86    \\
            MTPSL w/ reg. \cite{li2022learning} & 35.60 & 0.5576 & 29.70 & +19.66 \\
            \textbf{JTR w/o $\mathcal{L}_{Recon}$} & 35.89 & 0.5746 & 32.96 & +15.93 \\
            \textbf{JTR w/ $\mathcal{L}_{Recon}$} & {\bf 37.08} & {\bf 0.5541} & {\bf 29.44} & \textbf{+21.92} \\
            \midrule
            \multicolumn{5}{c}{\textit{Cityscapes ``onelabel''}}\\
            Supervised MTL & 69.50 & 0.0186 & --- & +0.000 \\
            MTPSL w/o reg. \cite{li2022learning} & 72.27 &0.0168 & --- & +6.832 \\
            MTPSL w/ reg. \cite{li2022learning} & 72.09 & 0.0168 & --- & +6.702 \\
            \textbf{JTR w/o $\mathcal{L}_{Recon}$} & 71.78 &  0.0166 & --- & +7.017 \\
            \textbf{JTR w/ $\mathcal{L}_{Recon}$} & \textbf{72.33} & {\bf 0.0163} & ---& \textbf{+8.219} \\
            \bottomrule
        \end{tabular}
    }
    \caption{Ablation study investigating the role of $\mathcal{L}_{Recon}$.
    }
    \vspace{-1em}
    \label{tab:abl-lrecon}
\end{table}

\paragraph{Reconstruction Loss}
We now present an ablation study on the role of the $\mathcal{L}_{Recon}$ term which prevents $g_{\theta_1}$ from learning a trivial encoding scheme (\ie mapping all inputs to a single point) in Tab. \ref{tab:abl-lrecon}. As expected, JTR performs more favorably with the inclusion of the $\mathcal{L}_{Recon}$ term. However, it is worth noting that JTR without $\mathcal{L}_{Recon}$ still outperforms both na\"ive Supervised MTL and MTPSL. This indicates that $g_{\theta_1}$ can manage to learn a useful joint-task space even without $\mathcal{L}_{Recon}$, and it is consistent with experiments from MTPSL \cite{li2022learning} which show that pair-wise task mapping still outperforms na\"ive Supervised MTL even if the mapping function is not explicitly regularized to prevent trivial solutions (also included in Tab. \ref{tab:abl-lrecon}). Given the additional computational costs of $\mathcal{L}_{Recon}$ (forward and backward passes through the decoder $g_{\theta_2}$) and the generally satisfactory performance of JTR without $\mathcal{L}_{Recon}$, omitting this term and relying solely on $g_{\theta_1}$ may be worth considering when under severe computational resource limitations.

\paragraph{Teacher-Student Pseudo-Labeling Baseline}
\label{sec:ablation-baseline}

In \Sref{sec:ssl-conreg}, we claim that consistency regularization is a reasonable choice for the baseline semi-supervised MTL method in our experiments, since combining single-task pseudo-labeling techniques for MTL is itself a major challenge~\cite{li2022learning}. However, for completeness, we also conduct an experiment with the MuST pseudo-labeling framework~\cite{ghiasi2021multi}.

MuST is a two-step approach which uses task-specific teacher models and a multi-task student model. When first training the $K$ teacher models, MuST uses only labeled samples and performs Single-Task Learning (STL). Then, a student model is trained with pseudo-labels generated by the teacher models for unlabeled tasks. Therefore, MuST performs poorly when labeled portions of the dataset are too small to train good teacher models for each task. This effect can be seen in the \Tref{tab:must}: both MTPSL and JTR outperform MuST because the underlying STL teacher models for MuST have subpar performance. Extrapolating from these results, we can infer that MuST's performance is likely less than the performance of MTPSL~\cite{li2022learning} in most benchmarks.

\begin{table}[t]
    \centering
    \tablescale{
    {
		\begin{tabular}{lcccc}
            \toprule
            Method & Seg. $\uparrow$ & Depth $\downarrow$ & Norm. $\downarrow$ & $\overbar{\Delta\%}$ $\uparrow$ \\
            \midrule
            Supervised MTL* & 27.05 & 0.6624 & 33.58 & +0.000 \\
            Consistency Reg.* & 29.50 & 0.6224 & 33.31 & +5.300 \\
            \midrule
            MuST STL Teachers~\cite{ghiasi2021multi} & 24.73 & 0.7278 & 29.78 & --2.378 \\
            MuST MTL Student~\cite{ghiasi2021multi} & 30.80 & 0.6185 & 31.82 & +8.577 \\
            \midrule
            MTPSL \cite{li2022learning}& 35.60 & 0.5576 & 29.70 & +19.66 \\
            \textbf{Ours (JTR)} & {\bf 37.08} & {\bf 0.5541} & {\bf 29.44} & \textbf{+21.92} \\
            \bottomrule
		\end{tabular}
    }
    }
    \caption{
        Benchmarks with MuST~\cite{ghiasi2021multi} on NYU-v2 ``randomlabel.'' Results marked * are directly quoted from \cite{li2022learning}.
    }
    \vspace{-1em}
    \label{tab:must}
\end{table}

\section{Discussion}

\paragraph{Limitations}

In this work, we investigate multi-task learning in the context of dense prediction tasks. 
However, applying JTR to MTL models for heterogeneous tasks, \ie a combination of instance-level classification, regression, and pixel-level dense prediction tasks, may not be straightforward. Extending JTR to arbitrary types of output domains would be an interesting future research direction.
Additionally, our experiments are conducted using single-domain data; however, in practical scenarios of MTL with partially labeled data, a more realistic and efficient approach would involve combining existing datasets from heterogeneous domains. This presents a greater challenge as the model needs to address both missing labels and domain gaps, but if successful, this may diversify the applicability of MTL.
Lastly, JTR does not take advantage of prior knowledge which can be manually modeled in some datasets. Remedying these limitations may bring further improvements.

\paragraph{Conclusion}
In this paper, we propose and demonstrate the effectiveness of applying Joint-Task Regularization (JTR) for multi-task learning with partially labeled data. JTR implicitly identifies cross-task relations through an encode-decode training process and uses this knowledge to regularize the MTL model in a joint-task space. Using variants of NYU-v2, Cityscapes, and Taskonomy, we comprehensively showcase the efficacy and efficiency of JTR in comparison to existing methods across a wide range of partially labeled MTL settings. We believe that JTR is a significant step forward in achieving data-efficient multi-task learning, and we hope that our contributions are helpful for future applications of MTL.

\paragraph{Acknowledgements} 
We thank all affiliates of the Harvard Visual Computing Group for their valuable feedback.
This work was supported by NIH grant R01HD104969.

{  
    \small
    \vspace{-2.5em}
    \bibliographystyle{ieeenat_fullname}
    \bibliography{main}
}

\clearpage
\setcounter{page}{1}
\maketitlesupplementary

\fakesection{Supplementary Material}

Here, we present additional material which could not be included in the main paper due to space constraints. 

\subsection{JTR Pseudo-Code}

\vspace{-1em}
\begin{algorithm}[h]
    \DontPrintSemicolon
    \small
    \For{$e = 1$ \KwTo $\mathrm{num\_epochs}$} {	
        From $\mathcal{D}$, draw a mini-batch $\mathcal{B} = ((x_0, y_0), \cdots, (x_B, y_B))$ \\
        \For{$(x, y) \in \mathcal{B}$} {
            \vskip 1em
            \tcp*[l]{get model predictions}
            $\hat{y} = h_{\psi}(f_{\phi}(x))$
            \vskip 1em
            \tcp*[l]{supervised loss for labeled tasks}
            $\mathcal{L} = 0$\\
            \For{$t \in \mathcal{T}_x$} {
                $ \mathcal{L} = \mathcal{L} + \frac{1}{|\mathcal{T}_x|} \mathcal{L}^t(\hat{y}^t, y^t)$
            }
            \vskip 1em
            \tcp*[l]{stack predictions and labels}
            $\hat{Y} = [\hat{y}^1_x \; \cdots \; \hat{y}^K_x]$\\
            $Y = [\delta(x, 1) \; \cdots \; \delta(x, K)]$\\
            \vskip 1em
            \tcp*[l]{apply distance loss}
            $\mathcal{L} = \mathcal{L} + c_{dist} \cdot \mathcal{L}_{Dist}(g_{{\theta_1}}(\hat{Y}), g_{\theta_1}(Y))$\\
            \tcp*[l]{apply reconstruction loss}
            $\mathcal{L} = \mathcal{L} + c_{recon} \cdot (\mathcal{L}_{Recon}(g(\langle {\hat{Y}} \rangle),\langle {\hat{Y}} \rangle) +\mathcal{L}_{Recon}(g(\langle {Y} \rangle), \langle{Y}\rangle))$\\
            \vskip 1em
            \tcp*[l]{update parameters}
            \For{$\zeta \in \{\phi, \psi, \theta_1, \theta_2\}$} 
            {
                $\zeta=\mathrm{SGD}(\mathcal{L},\zeta)$
            }
        }
    }
\caption{\small Pseudo-code for JTR. 
}
\end{algorithm}
\vspace{-1em}
In the above pseudo-code, $\delta(x, t) = \begin{cases} 
y^t_x & t \in \mathcal{T}_{x} \\
\hat{y}^t_x & t \notin \mathcal{T}_{x}
\end{cases}$, and $\langle\cdot\rangle$ detaches a variable from the graph to prevent backpropagation through the variable. 

\subsection{Implementation Details}
\label{app:impl}

\paragraph{Hyperparameters}

Across all experiments, we hold all hyperparameters constant for each method for all label scenarios. We use the default hyperparameter values provided by the official MTPSL source code repository for baseline methods \cite{li2022learning, mtpsl_gh}. For JTR, we keep hyperparameter adjustments to a minimum with the only new/modified parameters being: 
$c_{dist} = 4$,
$c_{recon} = 2$,
and the
batch size ($4$ for NYU-v2, $16$ for Cityscapes, $8$ for Taskonomy). For additional information, please refer to our code repository at \nohttpurl{github.com/KentoNishi/JTR-CVPR-2024}.

\paragraph{Auto-Encoder}
We use the same SegNet architecture as the model itself. The latent dimensions are $512\times9\times12$, and we did not tune this parameter. For an ablation study of this parameter, please see the ``Bottleneck Size'' ablation study in \Sref{sec:bottleneck-size}. In terms of interpretation, the JTR auto-encoder captures task relations which efficiently encode predictions and labels across multiple tasks in unison. The joint-task embeddings for $Y_x$ and $\hat{Y}_x$ can be thought of as compressed representations of predictions and targets in a shared space. The compression is facilitated by the reconstruction component of JTR---since the JTR encoder needs to compress its inputs to a reduced-dimension space, the encoder exploits commonalities and patterns across all tasks to create a compact and non-trivial encoding. This means that regularization applied in this feature space is not per-task; rather, all tasks are regularized at once.

\paragraph{Complexities of MTPSL}
\label{sec:complexities}
MTPSL relies on FiLM to condition on task-pairs. This involves adding parameters of shapes $(2C)$$\times K$$\times (K-1)$ and $1 \times C$ for \textit{every} Conv2D layer in MTPSL ($C$ channels, $K$ tasks), as well as implementing affine transformations for each layer. FiLM parameters also require a separate optimizer, LR, scheduler, etc. This makes using off-the-shelf architectures (\ie ResNet) challenging, since the model's inner code must be modified. In contrast, JTR has no per-layer intricacies and is easier to implement.

\paragraph{Batching}
JTR and all baselines use simple random batching for nearly all experiments. Two notable exceptions are NYU-v2 ``extended'' and ``Partially (Un)labeled MTL,'' for which we under/over-sample and append the additional data to each batch by a fixed quantity. The number of additional data examples added per batch are $4$ for NYU-v2 ``extended'' and $1$ for ``Partially (Un)labeled MTL.''

\subsection{Explaining Results}
\label{sec:explaining-results}

\begin{table}[t]
    \centering
    \tablescale{
    {
        \begin{tabular}{lcccl}
            \toprule
            Method & Seg. $\uparrow$ & Depth $\downarrow$ & Norm. $\downarrow$ & $\overbar{\Delta\%}$ $\uparrow$\\
            \midrule
            \multicolumn{5}{c}{\textit{NYU-v2 ``onelabel''}}\\
            Supervised MTL & 22.52 & 0.6920 & 35.27 & +0.000 \\
            MTPSL & 30.40 & 0.5926 & 31.68 &
            +19.84 \\
            \textbf{Ours (JTR)} & 31.96 & 0.5919 & 30.80 & 
            +23.02 \\
            \multicolumn{5}{c}{\textit{NYU-v2 ``randomlabels''}}\\
            Supervised MTL & 31.87 & 0.5957 & 31.64 & +0.000 \\
            MTPSL & 35.60 & 0.5576 & 29.70 & 
            +8.077 \\
            \textbf{Ours (JTR)} & 37.08 & 0.5541 & 29.44 & 
            +10.09 \\
            \multicolumn{5}{c}{\textit{Cityscapes ``onelabel''}}\\
            Supervised MTL & 68.35 & 0.0178 & --- & +0.000 \\
            MTPSL &  72.09 & 0.0168 & --- & 
            +5.545 \\
            \textbf{Ours (JTR)} & 72.33 & 0.0163 & --- & 
            +7.125 \\

            \bottomrule
        \end{tabular}
    }
    }
    \caption{Results for MTPSL and JTR relative to Supervised MTL results reproduced with RandAugment.}
    \label{tab:relative}
\end{table}

\paragraph{Copied Baselines}
For every experiment we ran, MTPSL and JTR use the same augmentation technique and dataloader. This ensures that any performance gain is solely due to JTR's advantages over MTPSL. However, for results for non-MTPSL baselines marked with * in some tables, we present scores directly sourced from\;\cite{li2022learning} to maintain consistency, since some baselines (\ie Direct / Perceptual Map) were not reproducible with RandAugment. For further clarity, we supplement our results with comparisons of MTPSL and JTR relative to Supervised MTL reproduced with RandAugment in \Tref{tab:relative}.

\paragraph{Statistical Significance}
MTL training is expensive; therefore, we followed previous works~\cite{li2022learning, liu2019end, standley2020tasks, yu2020gradient} in running each method only once using the same seed.
For additional clarity, we also ran JTR and MTPSL three times with the same seed on NYUv2 ``randomlabels.'' The 95\% confidence intervals in \Tref{tab:sd} show that JTR's $\overbar{\Delta\%}$ gap is statistically significant.

\begin{table}[t]
    \centering
    \tablescale{
    {
        \begin{tabular}{llccc}
            \toprule
            Dataset & Scenario & Seg. $\uparrow$ & Depth $\downarrow$ & Norm. $\downarrow$ \\
            \midrule
            NYU-v2 & \textit{onelabel} & 23.65 & 0.7513 & 30.56 \\
            NYU-v2 & \textit{randomlabels} & 28.68 & 0.7530 & 29.47 \\
            Cityscapes & \textit{onelabel} & 70.32 & 0.0134 & --- \\
            \bottomrule
        \end{tabular}
    }
    }
    \caption{Per-task Single-Task Learning (STL) results.}
    \label{tab:stl-benchmarks}
\end{table}

\begin{table}[t]
    \centering
    \tablescale{
    {
        \begin{tabular}{l|cc|c}
            \toprule
            & \multicolumn{2}{c|}{{NYU-v2}} & \multicolumn{1}{c}{{Cityscapes}}\\
            Method & \textit{onelabel} & \textit{randomlabels} & \textit{onelabel}\\
            \midrule
            Supervised MTL  & --4.100 & +8.216 & --17.82 \\
            MTPSL           & +15.33 & +16.43 & --11.43 \\
            \textbf{Ours (JTR)}      & +18.52 & +18.60 & --9.392 \\
            \bottomrule
        \end{tabular}
    }
    }
    \caption{$\overbar{\Delta\%}$ scores relative to per-task STL results on NYU-v2 and Cityscapes. All results were obtained with RandAugment.}
    \label{tab:stl-nyu-cityscapes}
\end{table}

\begin{table}[t]
    \centering
    \tablescale{
    {
        \begin{tabular}{lcccc}
            \toprule
            Method          & \textit{onelabel} & \textit{randomlabels} & \textit{halflabels} \\
            \midrule
            Supervised MTL  & +2.594     & --2.817 & --1.859 \\
            Consistency Reg.& +3.622     & +0.592 & +2.025 \\
            MTPSL           & +5.852     & +3.248 & +2.026 \\
            \textbf{Ours (JTR)}      & +5.482     & +3.919 & +2.785 \\
            \bottomrule
            
        \end{tabular}
        
    }
    }
    \caption{
        $\overbar{\Delta\%}$ scores relative to STL on Taskonomy.
    }
    \label{tab:stl-taskonomy}
    \vspace{-1em}
\end{table}

\paragraph{Single-Task Learning Baseline}
In our experiments, we report the $\overbar{\Delta\%}$ scores relative to \textit{Supervised MTL} because we believe it is the most relevant and fair comparison baseline. Alternatively, comparisons can be made against scores achieved with Single-Task Learning (STL). However, STL takes at least $N$-times more parameters, VRAM, and computation than an $N$-task MTL model, and is thus not a representative baseline. Nonetheless, we present comparisons with STL in \Tref{tab:stl-benchmarks}, \Tref{tab:stl-nyu-cityscapes}, and \Tref{tab:stl-taskonomy}.

\paragraph{Task Performance Tradeoffs}
In some experiments, JTR performs marginally lower than MTPSL in one task while scoring significantly higher in another (\Tref{tab:nyuv2}, \Tref{tab:taskonomy}, \Tref{tab:cityscapes-ssl}). Our understanding is that this is due to the tradeoff between tasks in MTL (\ie due to conflicting gradients\;\cite{yu2020gradient, Senushkin_2023_CVPR} and negative transfer\;\cite{liu2019loss, lakkapragada2023mitigating}). Crucially, JTR's lowest-scoring tasks are still competitive with MTPSL, while its highest-scoring tasks significantly outperform MTPSL, yielding better overall $\overbar{\Delta\%}$ performance.

\begin{table}[t]
    \centering
    \tablescale{
    {
        \setlength{\tabcolsep}{3pt}
        \begin{tabular}{lcccc}
            \toprule
            Method & Seg. $\uparrow$ & Depth $\downarrow$ & Norm. $\downarrow$ & $\overbar{\Delta\%}$ $\uparrow$ \\
            \midrule
            MTPSL & 35.45\plusminus0.247 & 0.5636\plusminus0.0059 & 29.95\plusminus0.344 & +18.92\plusminus0.88 \\
            JTR & 36.86\plusminus0.574 & 0.5517\plusminus0.0024 & 29.39\plusminus0.088 & +21.82\plusminus0.58 \\
            \bottomrule
        \end{tabular}
    }
    }
    \caption{Mean and SD on NYU-v2 \textit{randomlabels}. }
    \label{tab:sd}
\end{table}

\subsection{Ablation Studies}

\begin{table}[t]
    \centering
    \tablescale{{
		\begin{tabular}{lccccccccccc}
            \toprule
            Method & Seg. $\uparrow$ & Depth $\downarrow$ & Norm. $\downarrow$ & $\overbar{\Delta\%}$ $\uparrow$ \\
            \midrule
            Supervised MTL & 48.79 & 0.4528 & 25.35 & +0.000 \\
            Consistency Reg. & 46.02 & 0.4855 & 27.44 & --7.048 \\
            MTPSL \cite{li2022learning}  & 48.37 & 0.4228 & 25.39 & +1.869 \\
            \textbf{Ours (JTR)} & {\bf 49.07} & {\bf 0.4129} & {\bf 25.05} & \textbf{+3.523} \\
            \bottomrule
		\end{tabular}
    }}
    \caption{Partially labeled multi-task learning results on NYU-v2 ``randomlabels'' using DeepLabV3+ with ResNet-50.}
    \vspace{-1em}
    \label{tab:nyuv2-resnet}
\end{table}

\paragraph{Model-Agnosticism} Because JTR is a model-agnostic method, it can be applied to any dense multi-task prediction architecture. We demonstrate this general agnosticism by benchmarking ``Supervised MTL,'' ``Consistency Regularization,'', MTPSL \cite{li2022learning}, and JTR using a DeepLabV3+ model with the ResNet-50 architecture as the backbone. This study was carried out under the NYU-v2 ``randomlabels'' scenario using a batch size of $4$. 
We present the results in \Tref{tab:nyuv2-resnet}.
In comparison to the earlier SegNet experiments in~\Tref{tab:nyuv2}, all of the methods perform better; however, the model trained under JTR still obtains the best performance in all tasks, as well as the highest $\overbar{\Delta\%}$ score.  These results showcase the general applicability of JTR regardless of the model architecture. This setting corresponds to the ResNet-50 resource usage benchmarks in \Tref{tab:nyuperf} in the main paper.

\paragraph{Joint-Task Latent Distance Metric}
In our experiments, we primarily used cosine distance (dot product of normalized vectors) as our joint-task latent embedding distance metric.
However, other loss functions such as the L1/L2 distance can also be used as alternatives.
We present our comparisons between L1 norm, L2 norm, and cosine distance loss functions in \Tref{tab:loss}. These results show that cosine similarity is 
a favorable
distance metric for the joint-task latent space; hence, we use cosine similarity as $\mathcal{L}_{Dist}$ in all of our experiments.

\begin{table}[t]
    \centering
    \tablescale{
    {
		\begin{tabular}{lcccc}
            \toprule
            Function & Seg. $\uparrow$ & Depth $\downarrow$ & Norm. $\downarrow$ & $\overbar{\Delta\%}$ $\uparrow$ \\
            \midrule
            Supervised MTL & 31.87 & 0.5957 & 31.64 & +0.000 \\
            L1 Norm. & 35.21 & 0.5736 & 33.48 & +2.792  \\
            L2 Norm. & {35.11} & {0.5608} & {32.94} & +3.972  \\
            Cosine Distance &  {\bf 37.08} & {\bf 0.5541} & {\bf 29.44} & \textbf{+10.09} \\
            \bottomrule
		\end{tabular}
    }
    }
    \caption{
        Ablation study of loss functions for $\mathcal{L}_{Dist}$ with NYU-v2 ``randomlabels.'' All results were obtained with RandAugment.
    }
    \label{tab:loss}
\end{table}

\begin{table}[t]
    \centering
    \tablescale{
    {
		\begin{tabular}{lcccc}
            \toprule
            Dimensions & Seg. $\uparrow$ & Depth $\downarrow$ & Norm. $\downarrow$ & $\overbar{\Delta\%}$ $\uparrow$ \\
            \midrule
            Supervised MTL & 31.87 & 0.5957 & 31.64 & +0.000 \\
            $256\times9\times12$ & 35.93 & 0.5583 & 29.91 & +8.162  \\
            ${512\times9\times12}$ & {\bf 37.08} & {\bf 0.5541} & {\bf 29.44} & \textbf{+10.09} \\
            $1024\times9\times12$ & 35.69 & 0.5641 & 29.94 & +7.555 \\
            \bottomrule
		\end{tabular}
    }
    }
    \caption{
        Ablation study of the JTR auto-encoder bottleneck dimensions with NYU-v2 ``randomlabels.'' All results were obtained with RandAugment.
    }
    \vspace{-1em}
    \label{tab:bottleneck}
\end{table}

\paragraph{Bottleneck Size} 
\label{sec:bottleneck-size}
The latent dimensions of our SegNet auto-encoder in JTR are $512\times9\times12$ (the same dimensions as the MTL model). In \Tref{tab:bottleneck}, we perform an ablation study by varying the dimensions of the JTR auto-encoder bottleneck across $(256\times9\times12)$, $(512\times9\times12)$, and $(1024\times9\times12)$. We find that $(512\times9\times12)$ is the most effective bottleneck size. This is consistent with the dimensions of the SegNet instances used in the MTPSL codebase (for both the MTL model and the auxiliary modules)\;\cite{li2022learning, mtpsl_gh}.

\subsection{Additional Resource Usage Comparisons}

In \Tref{tab:nyuperf}, we present training time and VRAM usage comparisons for MTPSL \cite{li2022learning} and JTR. For completeness, we also benchmark the ``Supervised MTL'' and ``Consistency Regularization'' baselines in \Tref{tab:nyuperf-additional} and \Tref{tab:generalperf-additional}.
The results show that both MTPSL and JTR are more resource-intensive than na\"ive baselines. However, this tradeoff is worthwhile given the high prediction performance attained by the two methods in exchange. Furthermore, comparing the JTR training time to the supervised baseline on Taskonomy, JTR only takes $8$ additional hours to train. Meanwhile, MTPSL takes about $126$ additional hours. This shows that JTR indeed scales more favorably with more tasks.

\begin{table}[!t]
    \centering
    \tablescale{
    {
		\begin{tabular}{lrrrr}
            \toprule
            \multirow{2}{*}{Method} & \multicolumn{2}{c}{SegNet} & \multicolumn{2}{c}{ResNet-50} \\
            & Time $\downarrow$ & VRAM $\downarrow$ & Time $\downarrow$ & VRAM $\downarrow$\\
            \midrule
            Supervised MTL & 4h\:25m & 4GiB & 4h\:10m & 3.5GiB \\
            Consistency Reg. & 4h\:30m & 4GiB & 4h\:10m & 3.5GiB \\
            MTPSL \cite{li2022learning}& 8h\:20m & 17.5GiB & 17h\:40m & 21.5GiB\\
            \textbf{Ours (JTR)} & 9h\:00m & 17.5GiB & 10h\:50m & 16.2GiB \\
            \bottomrule
		\end{tabular}
    }
    }
    \caption{
        Time and VRAM requirements for training on NYU-v2 using a fixed batch size of $4$ on an NVIDIA A100 for $300$ epochs.
    }
    \label{tab:nyuperf-additional}
\end{table}

\begin{table}[!t]
    \centering
    \tablescale{
    {
		\begin{tabular}{lrrrr}
            \toprule
            \multirow{2}{*}{Method} & \multicolumn{2}{c}{Cityscapes} & \multicolumn{2}{c}{Taskonomy} \\
            & Time $\downarrow$ & VRAM $\downarrow$ & Time $\downarrow$ & VRAM $\downarrow$\\
            \midrule
            \multicolumn{5}{c}{\textit{Cityscapes (2 tasks), onelabel, SegNet}}\\
            Supervised MTL & 4h\:10m & 10.9GiB  & 97h\:00m  & 9.9GiB \\
            Consistency Reg. & 4h\:15m & 10.9GiB  & 97h\:00m & 10.2GiB \\
            MTPSL \cite{li2022learning}& 22h\:10m & 14.4GiB & 223h\:10m  &  34.4GiB \\
            \textbf{Ours (JTR)} & 23h\:45m & 19.2GiB  &  105h\:00m  &  23.9GiB \\
            \bottomrule
		\end{tabular}
    }
    }
    \caption{
        Time and VRAM requirements for training on Cityscapes and Taskonomy using the SegNet architecture on an NVIDIA A100, with $300$ and $20$ epochs respectively and batch sizes $16$ and $8$ respectively.
    }
    \vspace{-0.5em}
    \label{tab:generalperf-additional}
\end{table}

\section{Visualizations}
\begin{figure*}[t]
    \centering
    \tablescale{
    {
		\begin{tabular}{@{\hskip 7pt}c|@{\hskip 7pt}ccc|@{\hskip 7pt}ccc|@{\hskip 7pt}ccc}
            \toprule
            & \multicolumn{3}{c|@{\hskip 7pt}}{\hspace{-3.5pt}Segmentation} & \multicolumn{3}{c|@{\hskip 7pt}}{\hspace{-3.5pt}Depth} & \multicolumn{3}{c}{\hspace{-3.5pt}Normal} \\
            \hspace{-3.5pt}Input &\hspace{-3.5pt} \makebox[16pt][c]{MTPSL} &\hspace{-3.5pt} JTR &\hspace{-3.5pt} Label &\hspace{-3.5pt} \makebox[16pt][c]{MTPSL} &\hspace{-3.5pt} JTR &\hspace{-3.5pt} Label &\hspace{-3.5pt} \makebox[16pt][c]{MTPSL} &\hspace{-3.5pt} JTR &\hspace{-3.5pt} Label \\
            \midrule
            \tableimg{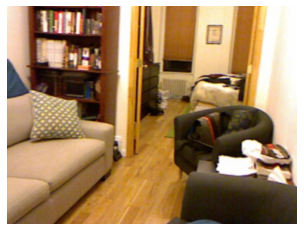} & \tableimg{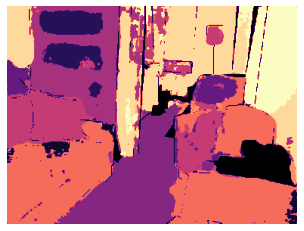} & \tableimg{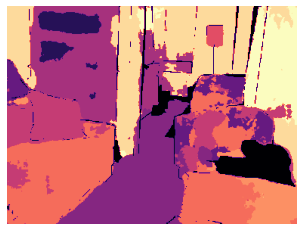} & \tableimg{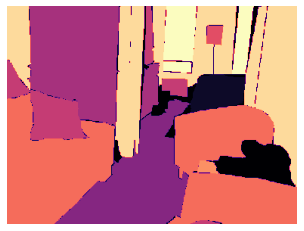} & \tableimg{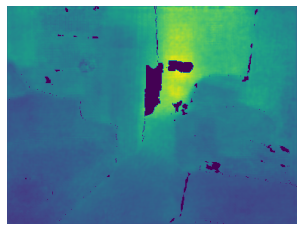} & \tableimg{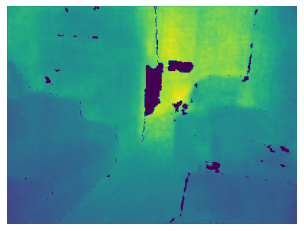} & \tableimg{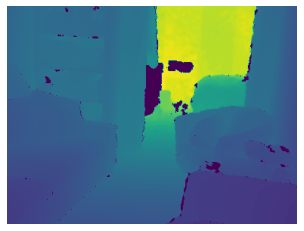} & \tableimg{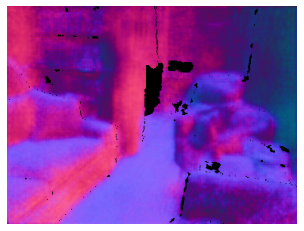} & \tableimg{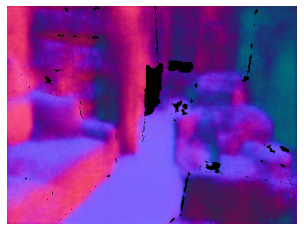} & \tableimg{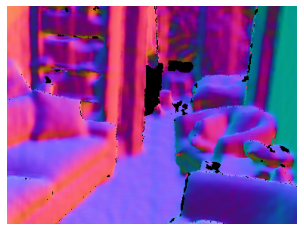} \\
            \tableimg{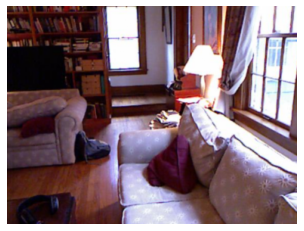} & \tableimg{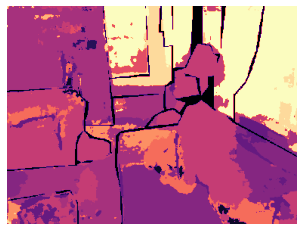} & \tableimg{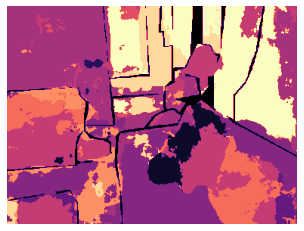} & \tableimg{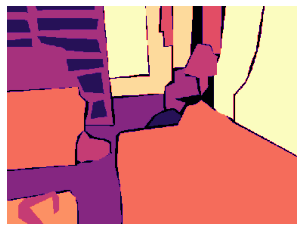} & \tableimg{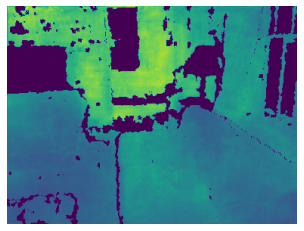} & \tableimg{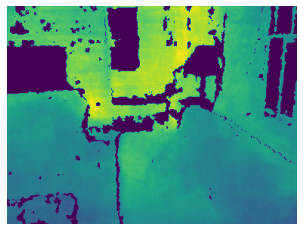} & \tableimg{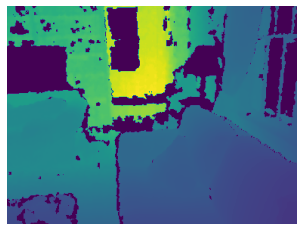} & \tableimg{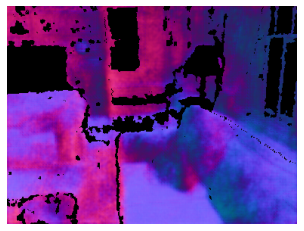} & \tableimg{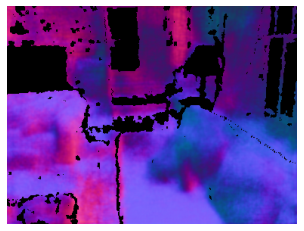} & \tableimg{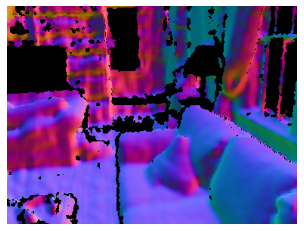} \\
            \tableimg{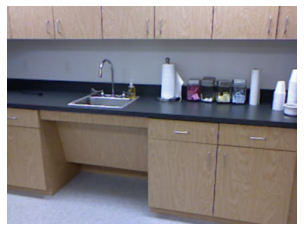} & \tableimg{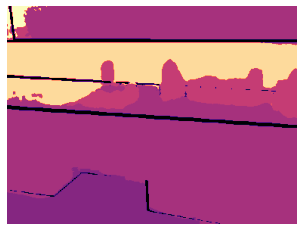} & \tableimg{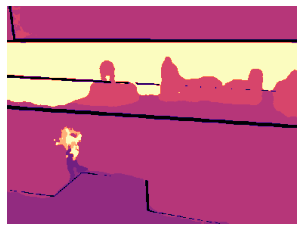} & \tableimg{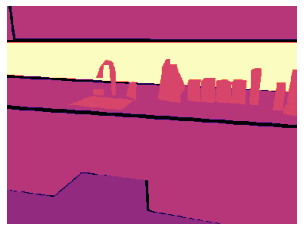} & \tableimg{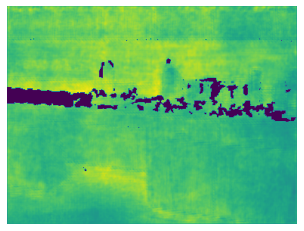} & \tableimg{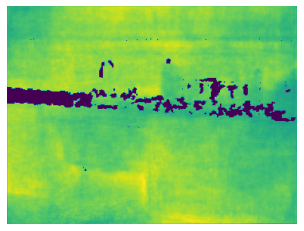} & \tableimg{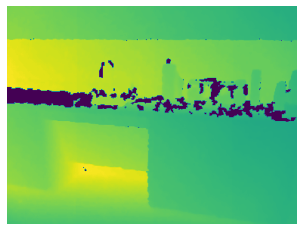} & \tableimg{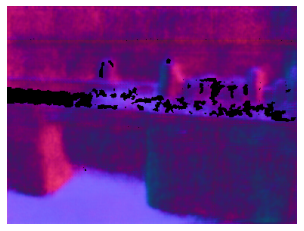} & \tableimg{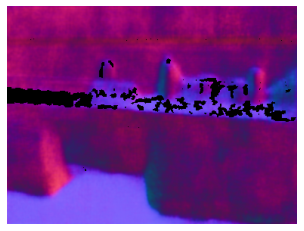} & \tableimg{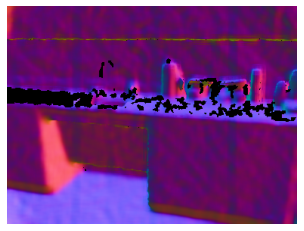} \\
            \tableimg{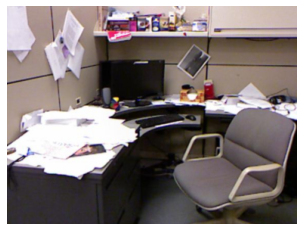} & \tableimg{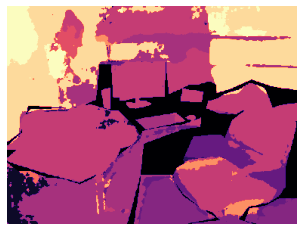} & \tableimg{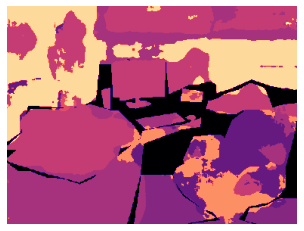} & \tableimg{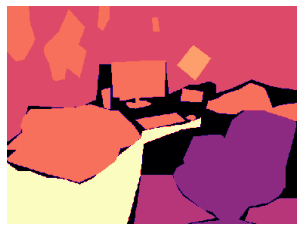} & \tableimg{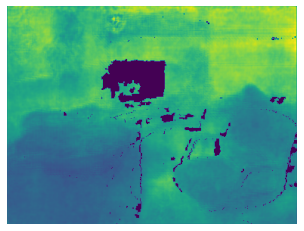} & \tableimg{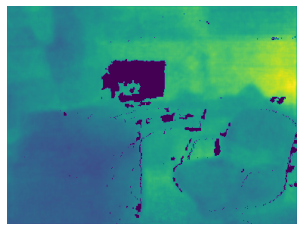} & \tableimg{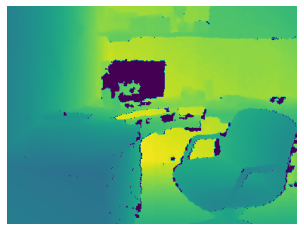} & \tableimg{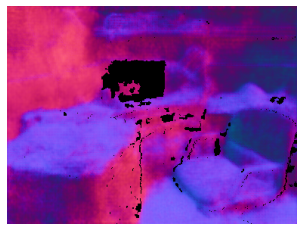} & \tableimg{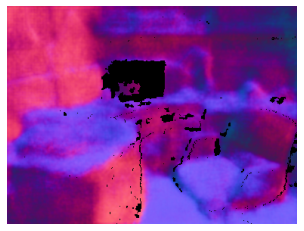} & \tableimg{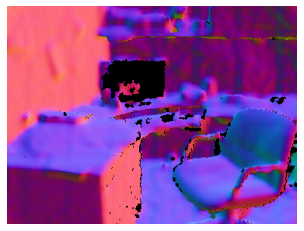} \\
            \tableimg{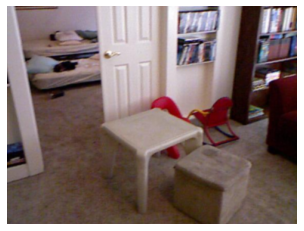} & \tableimg{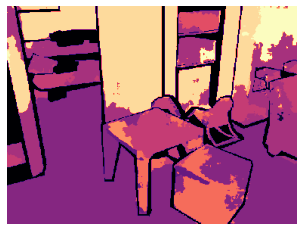} & \tableimg{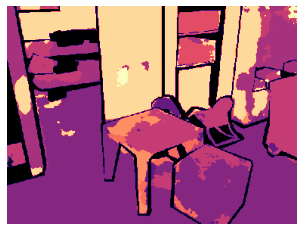} & \tableimg{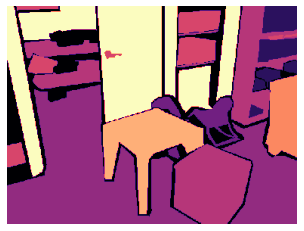} & \tableimg{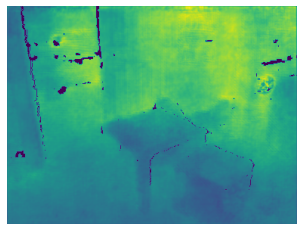} & \tableimg{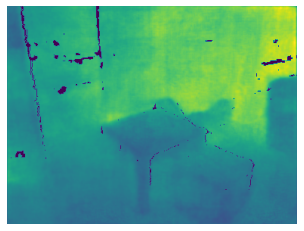} & \tableimg{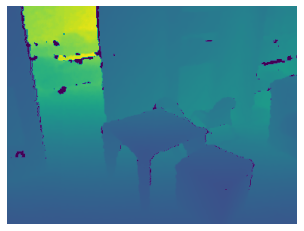} & \tableimg{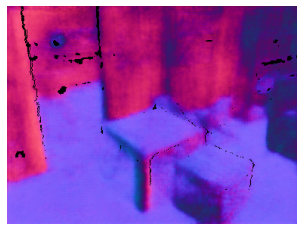} & \tableimg{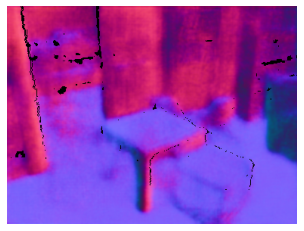} & \tableimg{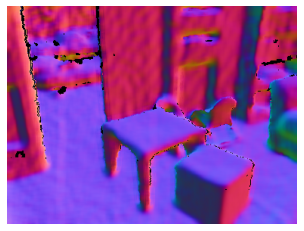} \\
            \tableimg{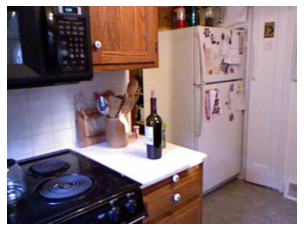} & \tableimg{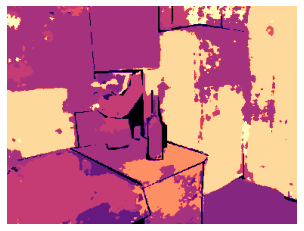} & \tableimg{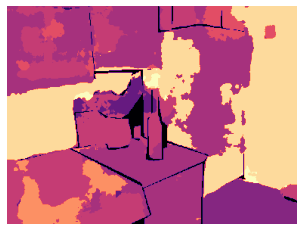} & \tableimg{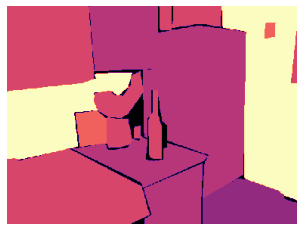} & \tableimg{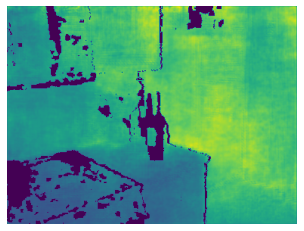} & \tableimg{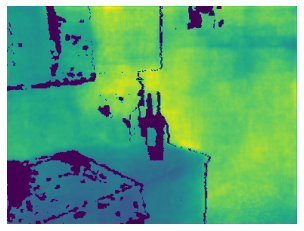} & \tableimg{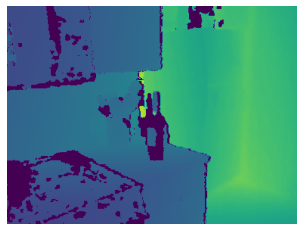} & \tableimg{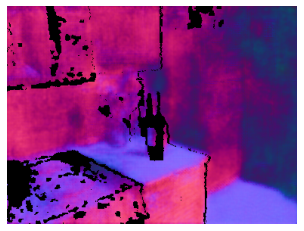} & \tableimg{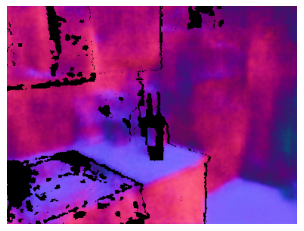} & \tableimg{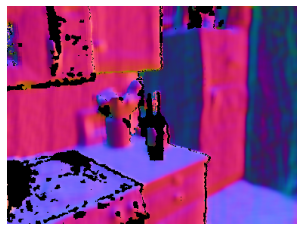} \\
            \tableimg{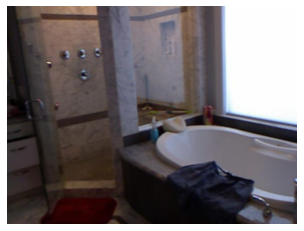} & \tableimg{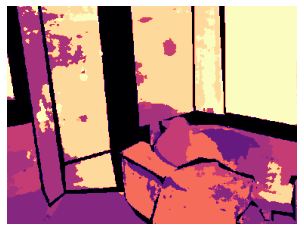} & \tableimg{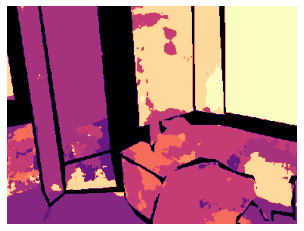} & \tableimg{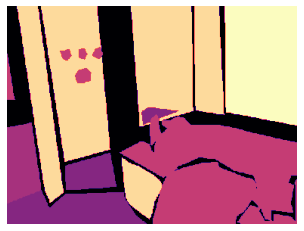} & \tableimg{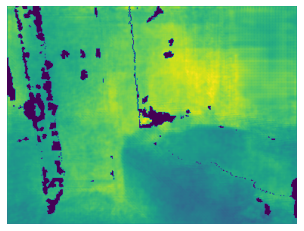} & \tableimg{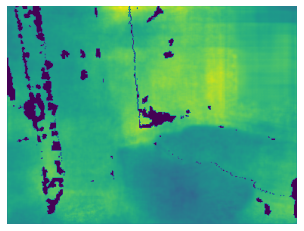} & \tableimg{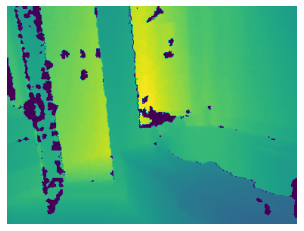} & \tableimg{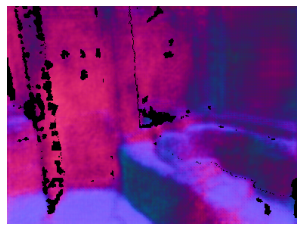} & \tableimg{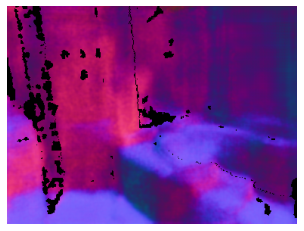} & \tableimg{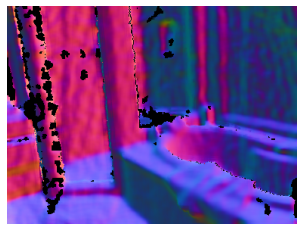} \\
            \tableimg{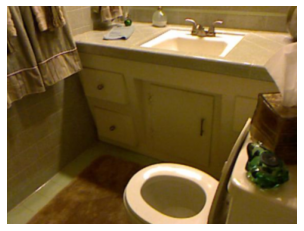} & \tableimg{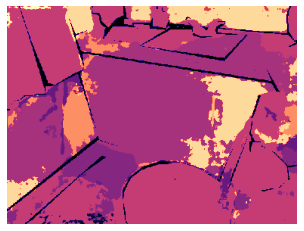} & \tableimg{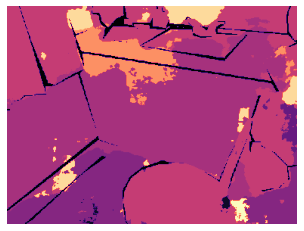} & \tableimg{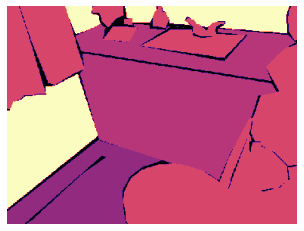} & \tableimg{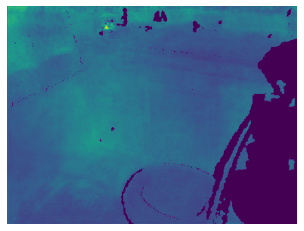} & \tableimg{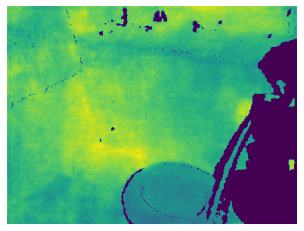} & \tableimg{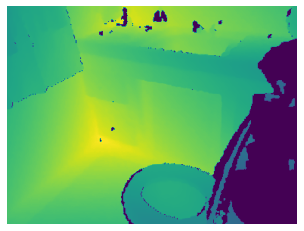} & \tableimg{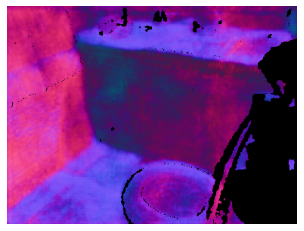} & \tableimg{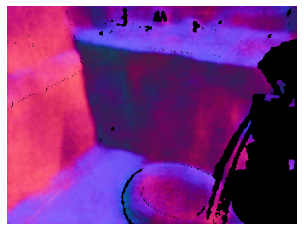} & \tableimg{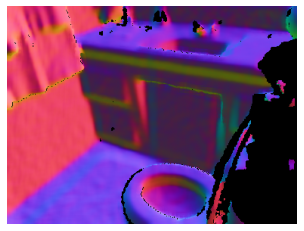} \\
            \tableimg{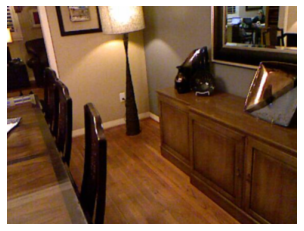} & \tableimg{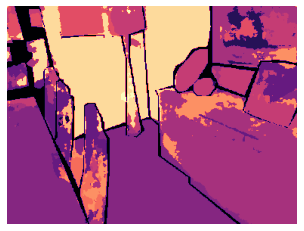} & \tableimg{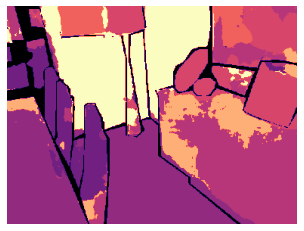} & \tableimg{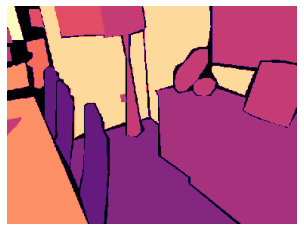} & \tableimg{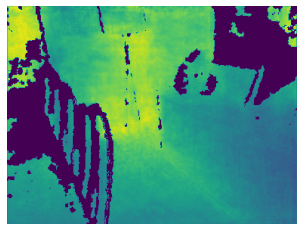} & \tableimg{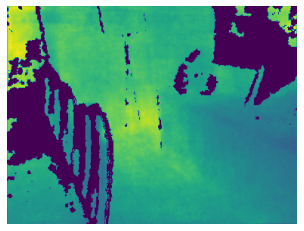} & \tableimg{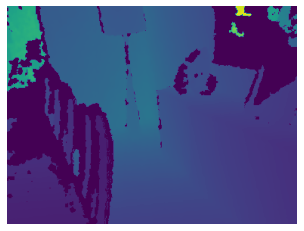} & \tableimg{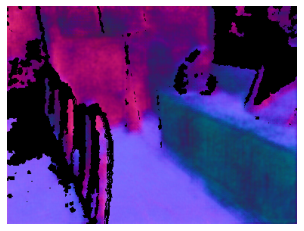} & \tableimg{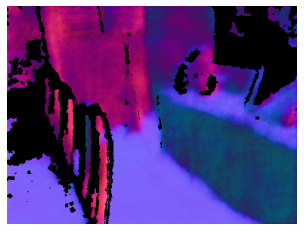} & \tableimg{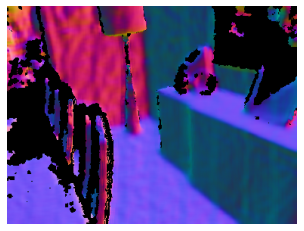} \\
            \tableimg{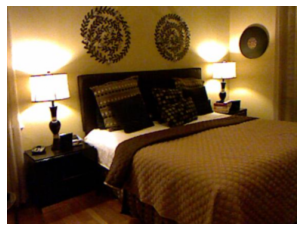} & \tableimg{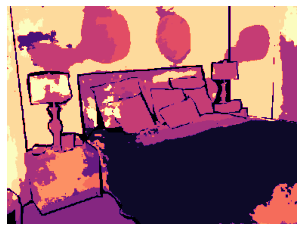} & \tableimg{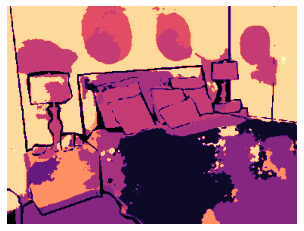} & \tableimg{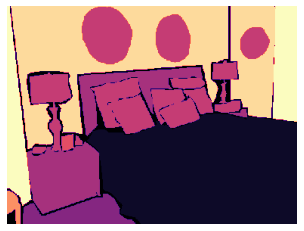} & \tableimg{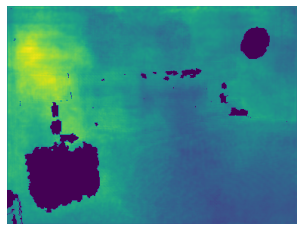} & \tableimg{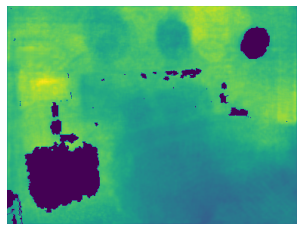} & \tableimg{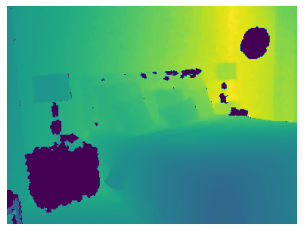} & \tableimg{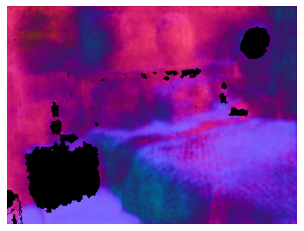} & \tableimg{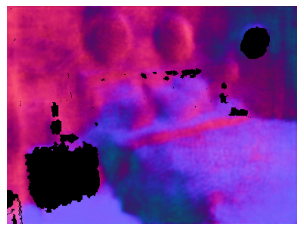} & \tableimg{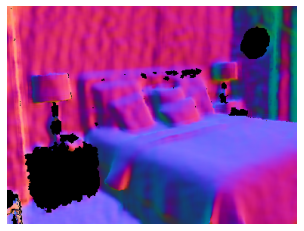} \\
            \bottomrule
		\end{tabular}
    }
    }
    \caption{
        Comparison of dense prediction outputs from a SegNet trained with MTPSL and JTR on NYU-v2 ``randomlabels'' alongside the corresponding input image and ground-truth labels. Examples are sampled randomly from the test set without cherry-picking. Visually, the predictions of the JTR model are slightly more accurate than those of the MTPSL model.
    }
    \label{tab:test-vis}
\end{figure*}

\begin{figure*}[t]
    \centering
    \tablescale{
    {
		\begin{tabular}{@{\hskip 7pt}c|@{\hskip 7pt}cccc|@{\hskip 7pt}cccc|@{\hskip 7pt}cccc@{\hskip 7pt}}
            \toprule
            \hspace{-3.5pt}Input & \multicolumn{4}{c|@{\hskip 7pt}}{\hspace{-7pt}Segmentation} & \multicolumn{4}{c|@{\hskip 7pt}}{\hspace{-7pt}Depth} & \multicolumn{4}{c}{\hspace{-7pt}Normal} \\
            \hspace{-3.5pt}$x$ &\hspace{-3.5pt} $\hat{y}_x$ &\hspace{-3.5pt} $g(\hat{Y}_x)$ &\hspace{-3.5pt} $y_x$ &\hspace{-3.5pt} $g(Y_x)$ &\hspace{-3.5pt} $\hat{y}_x$ &\hspace{-3.5pt} $g(\hat{Y}_x)$ &\hspace{-3.5pt} $y_x$ &\hspace{-3.5pt} $g(Y_x)$ &\hspace{-3.5pt} $\hat{y}_x$ &\hspace{-3.5pt} $g(\hat{Y}_x)$ &\hspace{-3.5pt} $y_x$ &\hspace{-3.5pt} $g(Y_x)$ \\
            \midrule
                        \tableimg{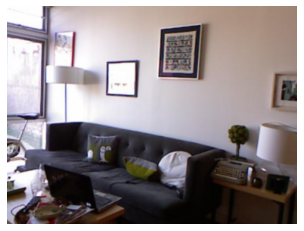} & \tableimg{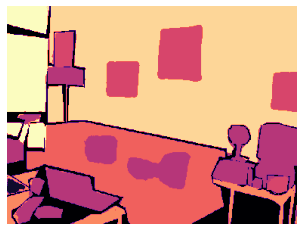} & \tableimg{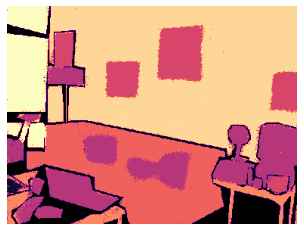} & \tableimg{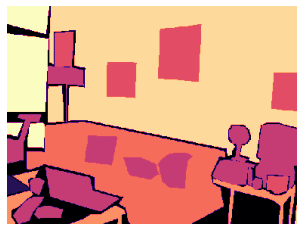} & \tableimg{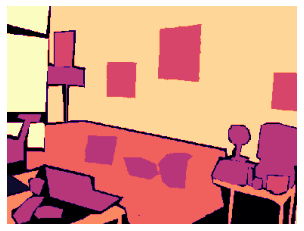} & \tableimg{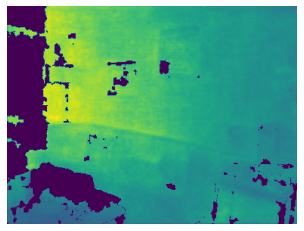} & \tableimg{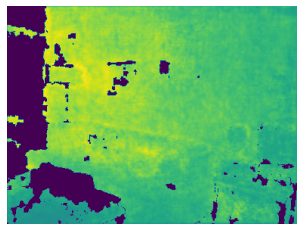} & \tableimg{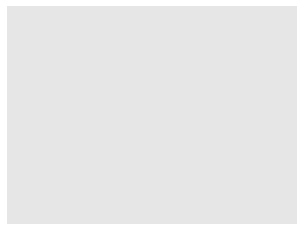} & \tableimg{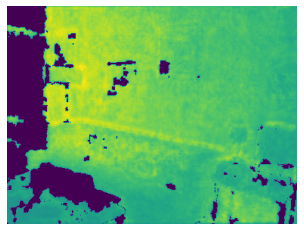} & \tableimg{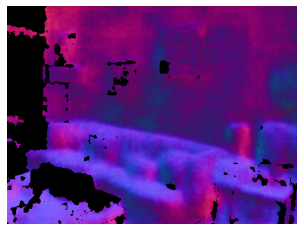} & \tableimg{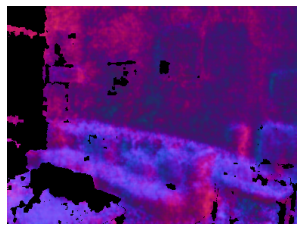} & \tableimg{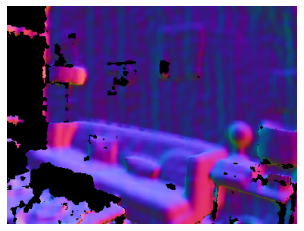} & \tableimg{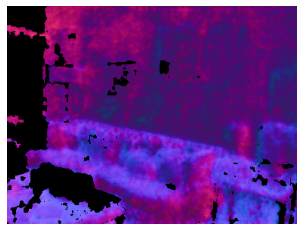}\\
                        \tableimg{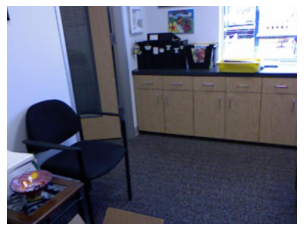} & \tableimg{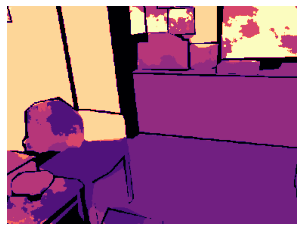} & \tableimg{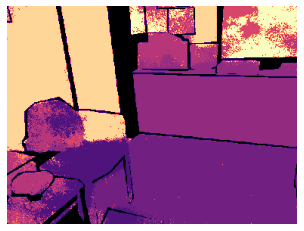} & \tableimg{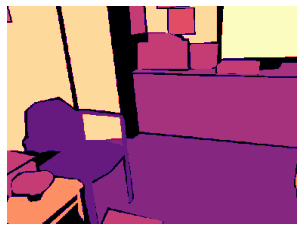} & \tableimg{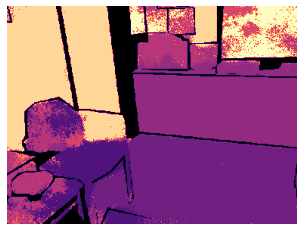} & \tableimg{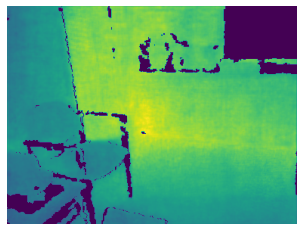} & \tableimg{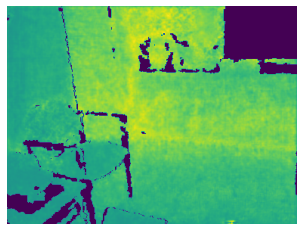} & \tableimg{figure/outputs/gray.png} & \tableimg{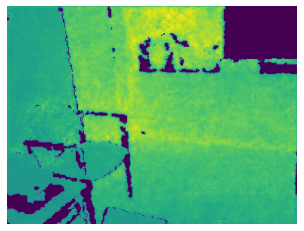} & \tableimg{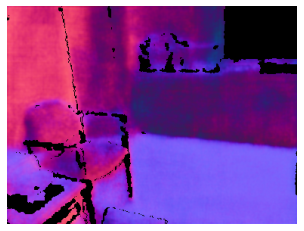} & \tableimg{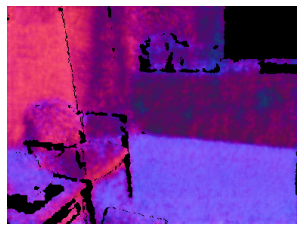} & \tableimg{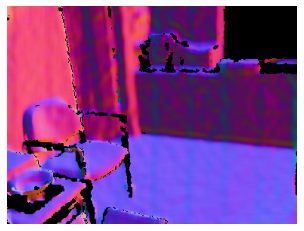} & \tableimg{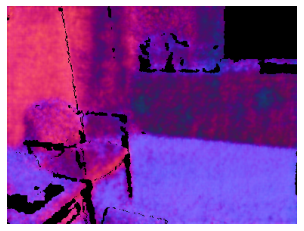}\\
                        \tableimg{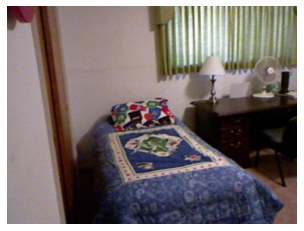} & \tableimg{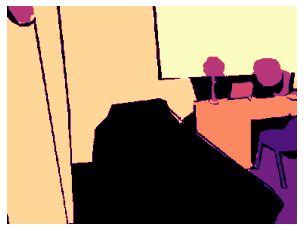} & \tableimg{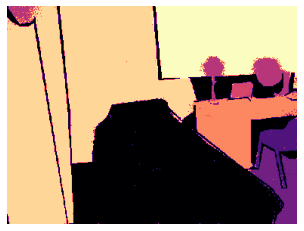} & \tableimg{figure/outputs/gray.png} & \tableimg{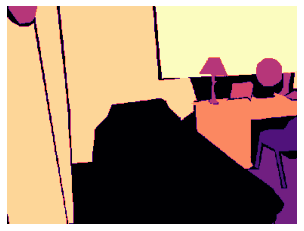} & \tableimg{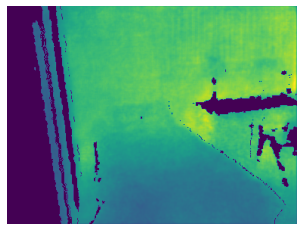} & \tableimg{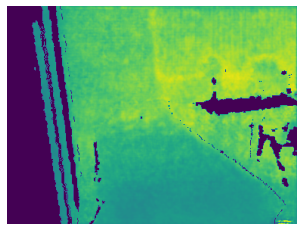} & \tableimg{figure/outputs/gray.png} & \tableimg{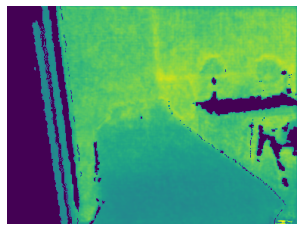} & \tableimg{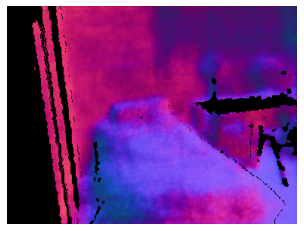} & \tableimg{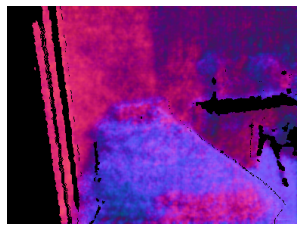} & \tableimg{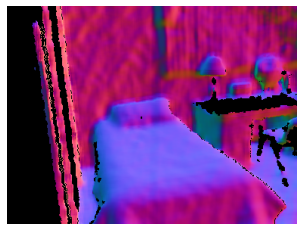} & \tableimg{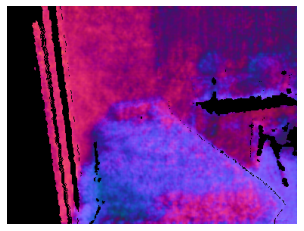}\\
                        \tableimg{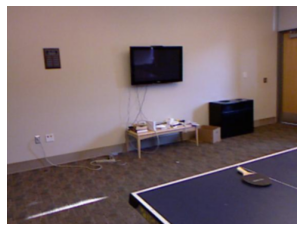} & \tableimg{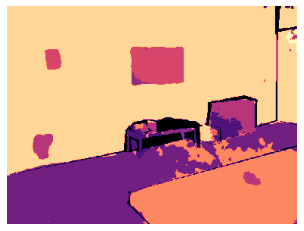} & \tableimg{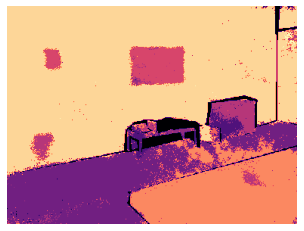} & \tableimg{figure/outputs/gray.png} & \tableimg{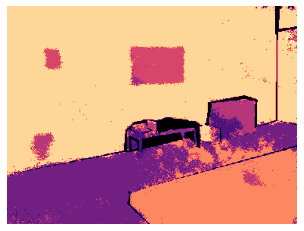} & \tableimg{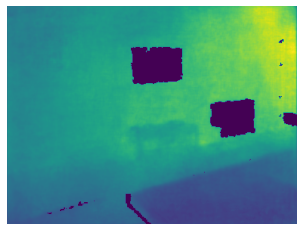} & \tableimg{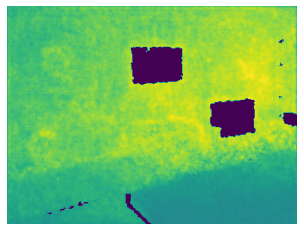} & \tableimg{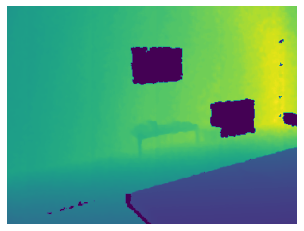} & \tableimg{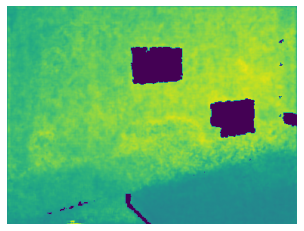} & \tableimg{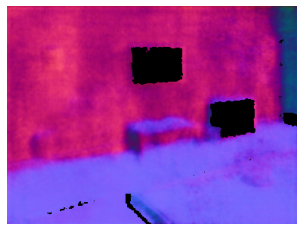} & \tableimg{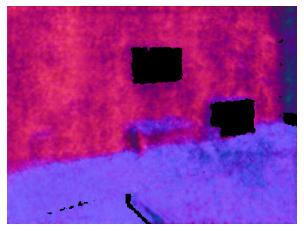} & \tableimg{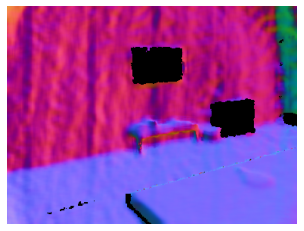} & \tableimg{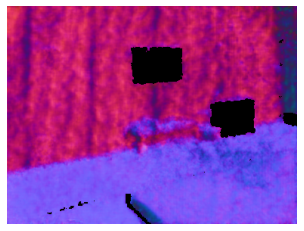}\\
                        \tableimg{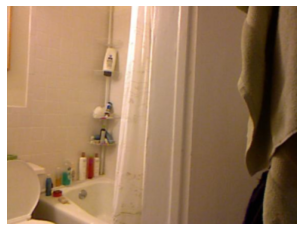} & \tableimg{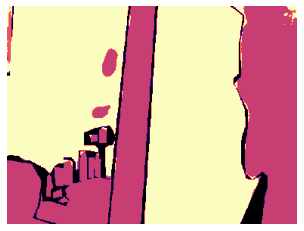} & \tableimg{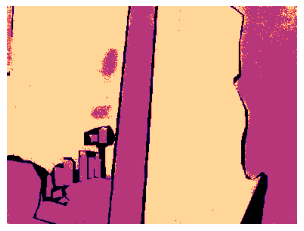} & \tableimg{figure/outputs/gray.png} & \tableimg{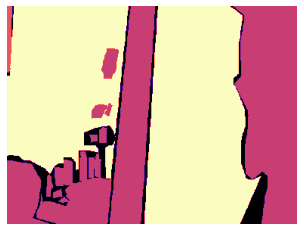} & \tableimg{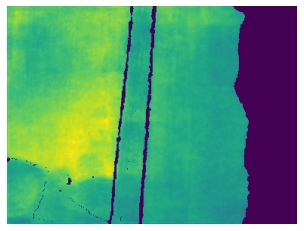} & \tableimg{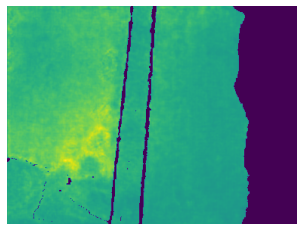} & \tableimg{figure/outputs/gray.png} & \tableimg{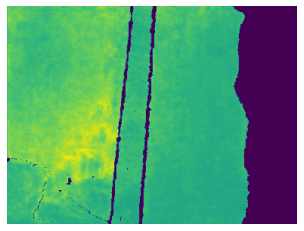} & \tableimg{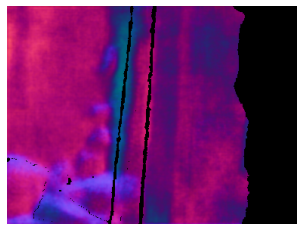} & \tableimg{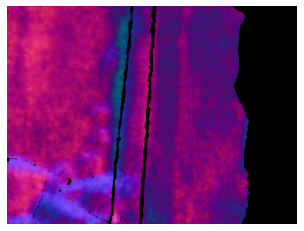} & \tableimg{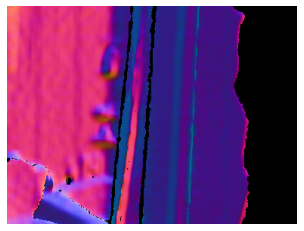} & \tableimg{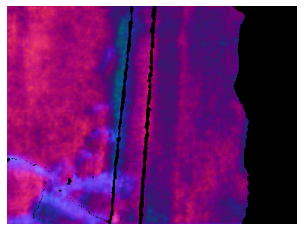}\\
                        \tableimg{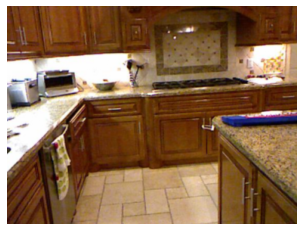} & \tableimg{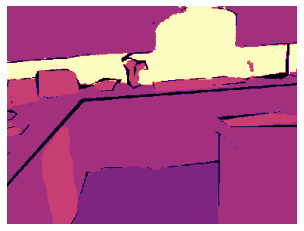} & \tableimg{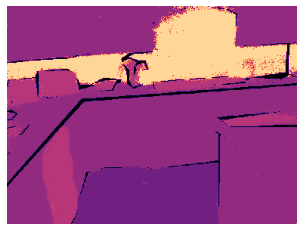} & \tableimg{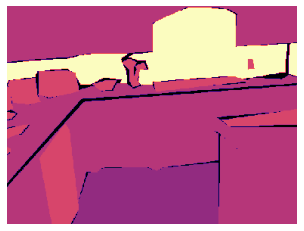} & \tableimg{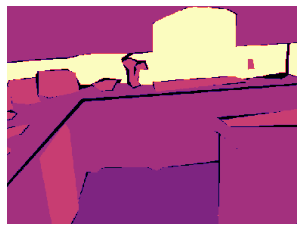} & \tableimg{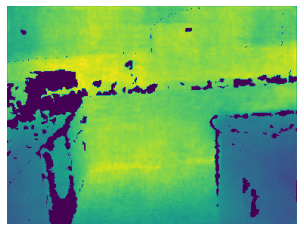} & \tableimg{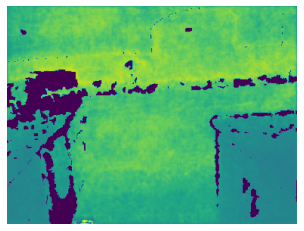} & \tableimg{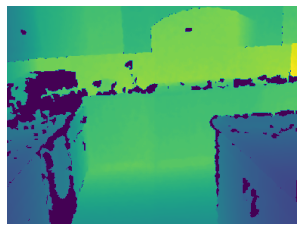} & \tableimg{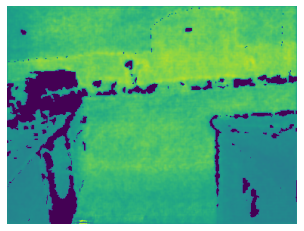} & \tableimg{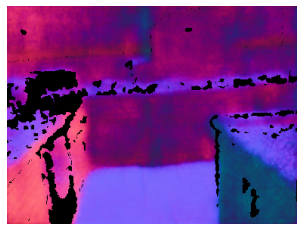} & \tableimg{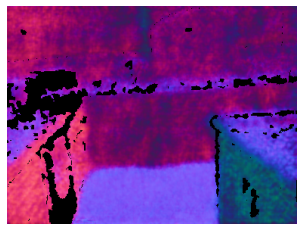} & \tableimg{figure/outputs/gray.png} & \tableimg{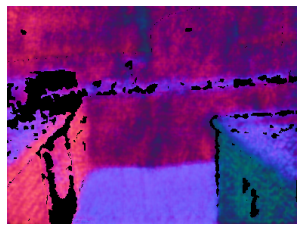}\\
                        \tableimg{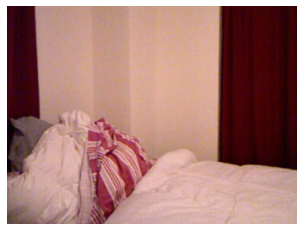} & \tableimg{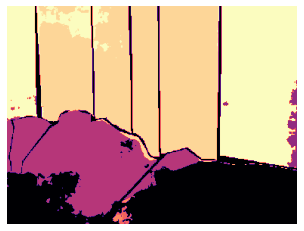} & \tableimg{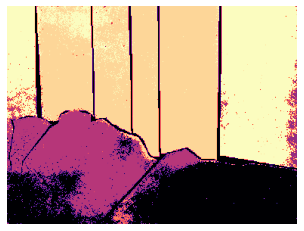} & \tableimg{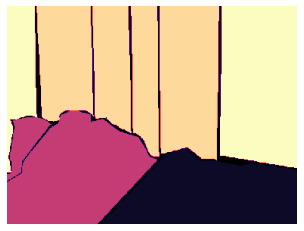} & \tableimg{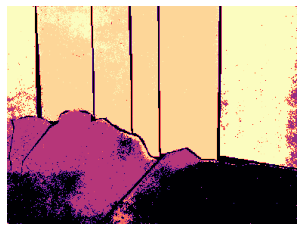} & \tableimg{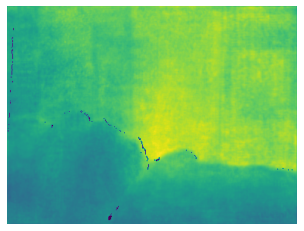} & \tableimg{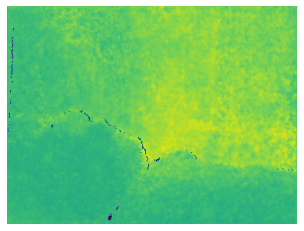} & \tableimg{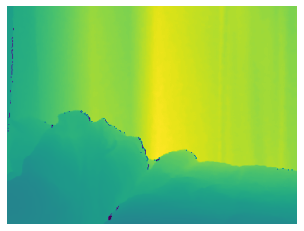} & \tableimg{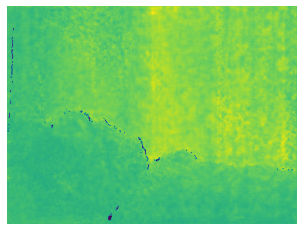} & \tableimg{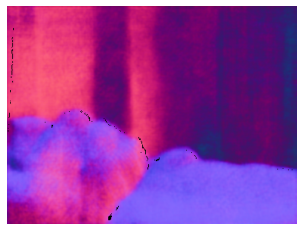} & \tableimg{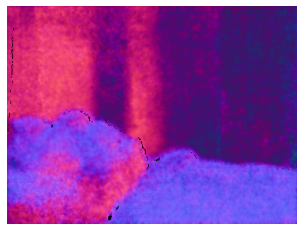} & \tableimg{figure/outputs/gray.png} & \tableimg{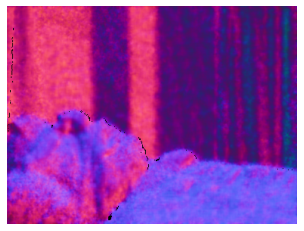}\\
                        \tableimg{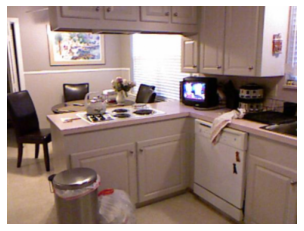} & \tableimg{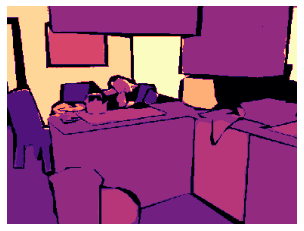} & \tableimg{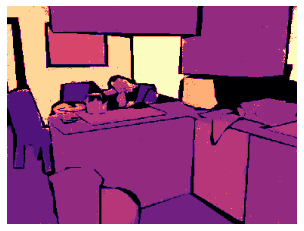} & \tableimg{figure/outputs/gray.png} & \tableimg{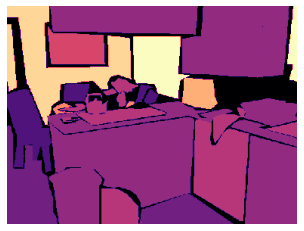} & \tableimg{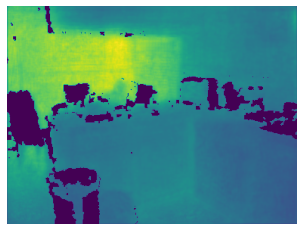} & \tableimg{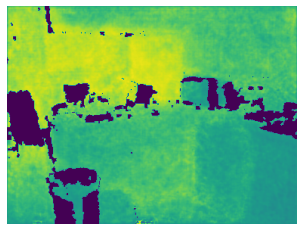} & \tableimg{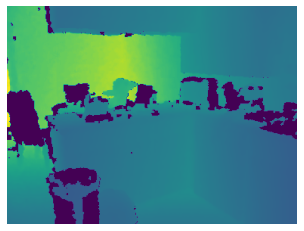} & \tableimg{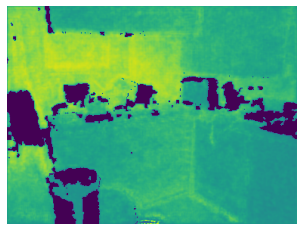} & \tableimg{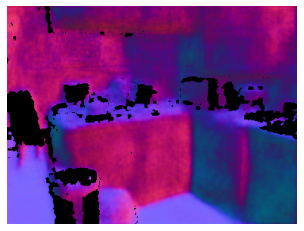} & \tableimg{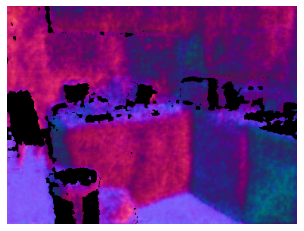} & \tableimg{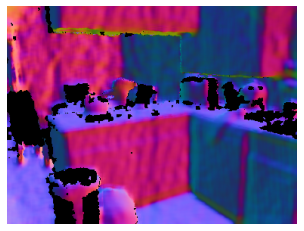} & \tableimg{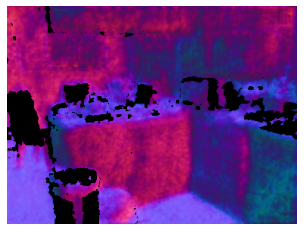}\\
                        \tableimg{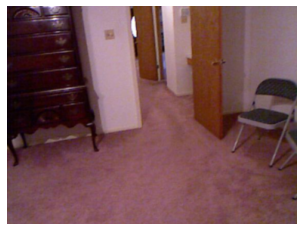} & \tableimg{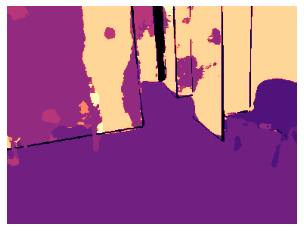} & \tableimg{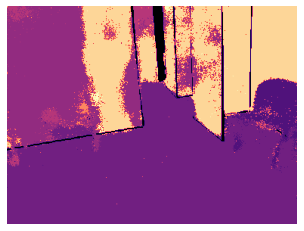} & \tableimg{figure/outputs/gray.png} & \tableimg{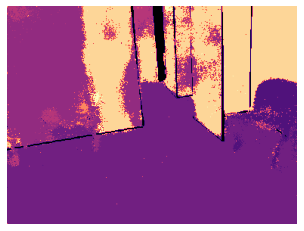} & \tableimg{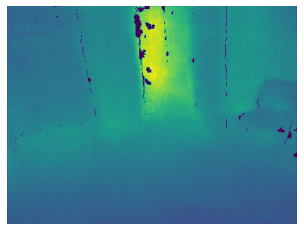} & \tableimg{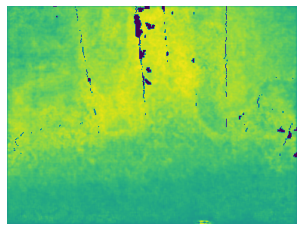} & \tableimg{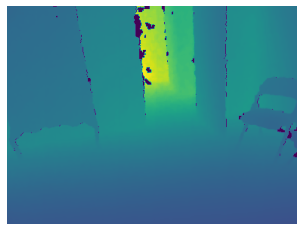} & \tableimg{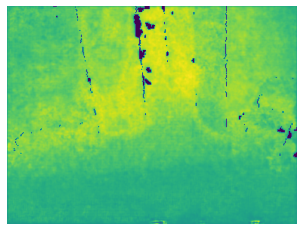} & \tableimg{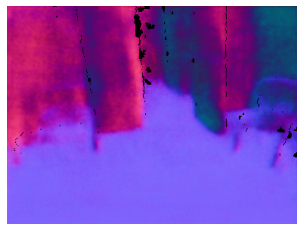} & \tableimg{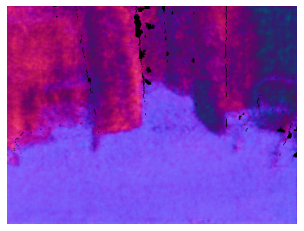} & \tableimg{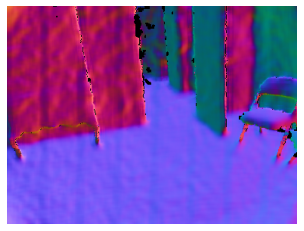} & \tableimg{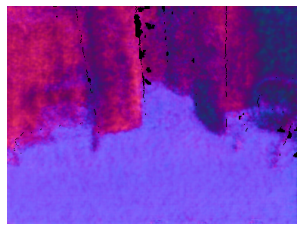}\\
                        \tableimg{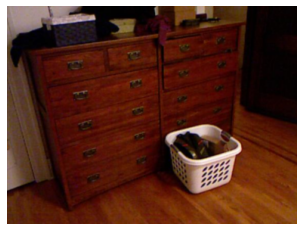} & \tableimg{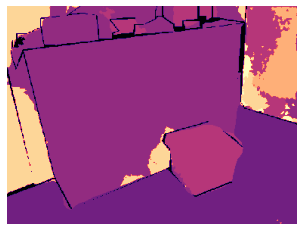} & \tableimg{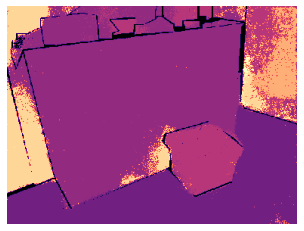} & \tableimg{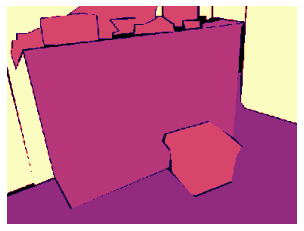} & \tableimg{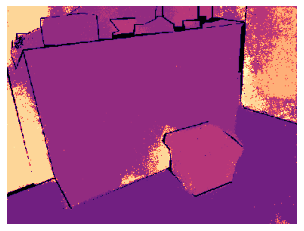} & \tableimg{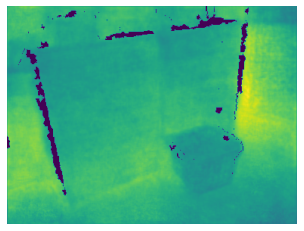} & \tableimg{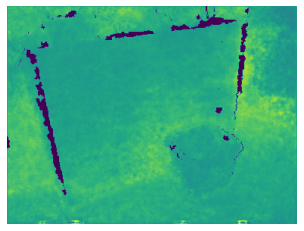} & \tableimg{figure/outputs/gray.png} & \tableimg{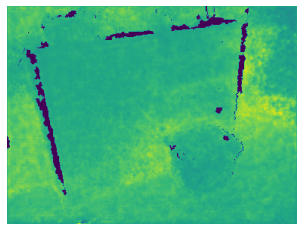} & \tableimg{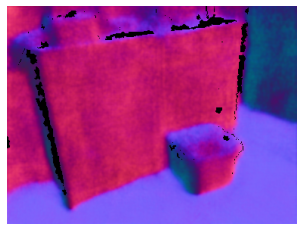} & \tableimg{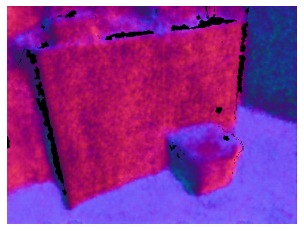} & \tableimg{figure/outputs/gray.png} & \tableimg{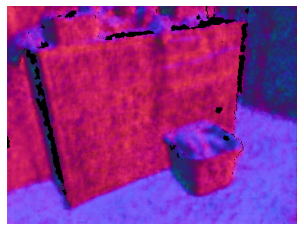}\\

            \bottomrule
		\end{tabular}
    }
    }
    \caption{
        Visualization of input images $x$, dense prediction outputs $\hat{y}$, labels $y$, JTR reconstructions of the noisy prediction tensor $g(\hat{Y}_x)$, and JTR reconstructions of the reliable target tensor $g(Y_x)$ for a SegNet trained with JTR on NYU-v2 ``randomlabels.'' Examples are randomly sampled from the training set without cherry-picking. Missing $y_x$ entries indicate unlabeled tasks under ``randomlabels.''
    }
    \label{tab:train-vis}
\end{figure*}

We provide comparisons of model predictions and labels for examples in the NYU-v2 test set in \Fref{tab:test-vis}. Additionally, we provide visualizations of JTR auto-encoder inputs and reconstructions for examples in the NYU-v2 ``randomlabels'' training set in \Fref{tab:train-vis}.

\end{document}